\documentclass[10pt]{article} 

\usepackage[preprint]{tmlr}

\usepackage{hyperref}
\usepackage{url}

\author{\name Mikael von Strauss \email mvonstra@amd.com \\
      \addr AMD Silo AI\\
      Stockholm, Sweden}
      
\def\month{11}  
\def\year{2025} 
\def\openreview{\url{https://openreview.net/forum?id=XXXX}} 

\usepackage{amsmath,amssymb,amsthm,bbold,braket}
\usepackage{graphicx}
\usepackage{subcaption}
\usepackage{textcomp}
\usepackage{enumitem}
\usepackage{import}

\hypersetup{
  colorlinks = true,
  urlcolor = blue,  
  pdfauthor = {Mikael von Strauss},
  pdfkeywords = {transformer, invertibility}, 
  pdftitle = {tigr},
  pdfsubject = {transformer_invertibility}
}

\newtheorem{theorem}{Theorem}[section]

\newtheorem{lem}[theorem]{Lemma}
\newtheorem{cor}[theorem]{Corollary}

\theoremstyle{definition} 
\newtheorem{defn}[theorem]{Definition}

\title{Transformer Injectivity \& Geometric Robustness\\[0.5em]
  \large\textit{Analytic Margins and Bi-Lipschitz Uniformity of Sequence-Level Hidden States}}

\begin{document}

\maketitle

\begin{abstract}
Under real-analytic assumptions on decoder-only Transformers, recent work shows that the map from discrete prompts to last-token hidden states is generically injective on finite prompt sets. We refine this picture: for each layer $\ell$ we define a collision discriminant $\Delta^\ell \subset \Theta$ and injective stratum $\mathcal{U}^\ell = \Theta \setminus \Delta^\ell$, and prove a dichotomy---either the model is nowhere injective on the set, or $\mathcal{U}^\ell$ is open and dense and every $F^\ell_\theta$ is injective. Under mild non-singularity assumptions on the optimizer and an absolutely continuous initialization, generic injectivity persists along smooth training trajectories over any fixed horizon. We also treat symmetry groups $G$, showing that discriminants and injective strata descend to the quotient $\Theta/G$, so injectivity is naturally a property of functional equivalence classes.

We complement these results with an empirical study of layerwise geometric diagnostics. We define a separation margin and a co-Lipschitz (lower Lipschitz) constant between prompt space and last-token representation space, estimated via nearest-neighbor statistics on large prompt sets. Applying these diagnostics to pretrained LLaMA-3 and Qwen models, we study behavior across layers, sequence lengths, model scales, and 8- and 4-bit activation quantization. On our sampled prompts we see no collisions in full precision or at 8 bits, while 4-bit quantization induces a small number of collisions and markedly shrinks co-Lipschitz estimates. For a small GPT-2 trained from scratch, normalized metrics remain stable over training. Overall, the results suggest that Transformer representations are generically and persistently injective in the continuous-parameter idealization, while their practical invertibility can be probed using simple geometric diagnostics.
\end{abstract}

\section{Introduction}

The question of whether large language models (LLMs) are ``invertible''—whether one can recover an input prompt from an internal hidden state—has attracted increasing interest, both for mechanistic interpretability and for privacy and memorization concerns. From a purely topological perspective, injectivity is not obviously ruled out: for a fixed context length and vocabulary, the prompt set is finite, while hidden states live in a high-dimensional continuous space. We emphasize that this finiteness is crucial: our injectivity results concern the finite set of token sequences up to a fixed context length, not the full combinatorial space of all possible sequences. Any obstruction to invertibility is therefore geometric and measure-theoretic rather than purely combinatorial. Beyond revisiting this finiteness-based perspective, our goal is to turn the existing “almost sure injectivity” results into a more detailed, layerwise and trajectory-aware picture, and to connect that picture to concrete geometric diagnostics that can be computed on modern pretrained LLMs.

Recent work by \cite{original_paper} showed that, under real-analytic assumptions on the building blocks (smooth activations, LayerNorm with $\epsilon>0$, softmax, causal attention, etc.), the forward map from prompts to last-token hidden states in decoder-only Transformers is generically injective: the set of parameter values that induce collisions is an analytic subset of Lebesgue measure zero. In particular, for almost all parameters, distinct prompts in the discrete prompt set produce distinct last-token representations at the final layer.

We begin by revisiting and slightly reorganizing this analytic picture. Using collision sets and their union as a discriminant $\Delta^\ell$ at each layer $\ell$, we define an injective stratum $\mathcal{U}^\ell = \Theta \setminus \Delta^\ell$ in parameter space and show that, for each layer separately, either the model is nowhere injective, or $\mathcal{U}^\ell$ is open and dense and every map $F^\ell_\theta$ with $\theta \in \mathcal{U}^\ell$ is injective on the finite prompt set. We then prove a simple persistence result: if the initialization has an absolutely continuous law and the optimizer induces non-singular, absolute-continuity-preserving updates (as in smooth GD/SGD/Adam in the usual real-valued idealization), then with probability one the parameter trajectory remains in the injective stratum at all layers up to any fixed training horizon. We also extend the framework to architectures with an internal symmetry group $G$ (e.g., permutations and compatible rescalings of neurons or heads), and show that the discriminant and injective strata descend to the quotient $\Theta/G$, so that generic injectivity is naturally a property of functional equivalence classes rather than raw weights.

The main contribution of this paper is empirical. Motivated by the theoretical background above, we introduce layerwise geometric diagnostics on large finite prompt sets: (i) a minimal separation margin, capturing how far apart the hardest-to-separate prompts are in representation space, and (ii) a co-Lipschitz (lower Lipschitz) constant between prompt space (equipped with Hamming distance) and last-token representation space, capturing how much the most contractive pairs of representations move per unit of Hamming distance. From these quantities we derive practical estimators based on nearest-neighbor statistics and worst-percentile ratios. We apply these diagnostics to a range of pretrained decoder-only models from the LLaMA-3 and Qwen families, studying how separation behaves across layers, as a function of sequence length, under model scaling, and under non-analytic post-hoc distortions such as 8-bit and 4-bit activation quantization. On large finite prompt sets we observe no exact collisions in any of the models we test, in line with generic injectivity, but worst-percentile co-Lipschitz constants can shrink substantially with longer contexts and under aggressive quantization. Finally, by pretraining a small GPT-2 model from scratch, we study how these metrics evolve during training and observe no systematic drift.

Analytically, in the continuous-parameter idealization, Transformers are generically injective and remain so under typical training dynamics of smooth non-singular updates; empirically, they behave like smooth, roughly scale-invariant embeddings of the prompt set. Our layerwise margins and co-Lipschitz constants provide a concrete notion of ``how invertible'' a given model is in practice—one that is stable across models, depths, and context lengths, yet sensitive to non-analytic distortions.

\paragraph{Summary of empirical findings.}
Across all models and datasets we study, we observe the following recurring patterns.
\begin{enumerate}[label=(\roman*)]
\item On large finite prompt sets, we do not encounter any exact collisions in last-token representations at any layer of the LLaMA-3 or Qwen models in full precision, nor under post-hoc 8-bit activation quantization, consistent with generic injectivity. Under more aggressive 4-bit activation quantization we do observe a modest number of collisions, concentrated in deeper layers, reflecting the non-injectivity of the quantizer itself.
\item Raw separation margins and co-Lipschitz estimates tend to grow with depth due to norm inflation, but their normalized counterparts (after rescaling by a simple per-layer norm factor) remain roughly stable across layers and model scales.
\item Worst-percentile co-Lipschitz constants systematically shrink with increasing context length and under increasingly aggressive post-hoc activation quantization, indicating more contractive behavior and a closer approach to practical non-invertibility on hard prompt pairs.
\item When we compare models across architectural families, normalized layerwise profiles of margins and co-Lipschitz constants cluster tightly within each family (e.g.,\ LLaMA versus Qwen) and are separated by a roughly constant offset between families. Our diagnostics thus act as a geometric “fingerprint’’ that is sensitive to architecture and training recipe rather than just parameter count.
\item Along the pretraining trajectory of a small GPT-2 model, these metrics exhibit no systematic drift: both normalized margins and co-Lipschitz estimates fluctuate within a narrow band, suggesting that typical training dynamics preserve the “distance to collision’’ established at initialization.
\end{enumerate}

\paragraph{Related work.}
Our work intersects two active lines of research: analytic studies of Transformer architectures and inverse problems in language modeling.

On the analytic side, several works have highlighted non-injectivity phenomena when Transformers are viewed as maps on continuous spaces. LayerNorm collapses along per-example statistics \citep{ba2016layernormalization}, residual connections can cancel, attention-only stacks can exhibit doubly-exponential rank decay with depth \citep{dong2023attentionneedpureattention}, and the softmax bottleneck constrains the set of realizable output distributions \citep{yang2018breakingsoftmaxbottleneckhighrank}. In parallel, a separate line of work designs architectures that are exactly invertible by construction, such as normalizing flows and reversible residual networks \citep{dinh2017densityestimationusingreal, kingma2018glowgenerativeflowinvertible, gomez2017reversibleresidualnetworkbackpropagation}, ensuring bijectivity at the architectural level rather than as a generic property. More recently, analytic and measure-theoretic viewpoints have been used to obtain “almost sure’’ structural guarantees: \citep{jiang2025surjectivityneuralnetworkselicit} show that common building blocks are generically surjective, while \citet{sutter2025nonlinearrepresentationdilemmacausal} prove that Transformers at random initialization are almost surely injective with respect to the full hidden-state matrix (but only at initialization, without training dynamics). We build directly on \citet{original_paper}, who establish generic injectivity of decoder-only Transformers on the discrete prompt set under real-analytic assumptions. Our analytic contribution is to (i) extend their result to a layerwise collision discriminant and injective stratum for last-token states, (ii) prove a persistence theorem showing that, under mild non-singularity assumptions on the optimizer and an absolutely continuous initialization, generic injectivity is almost surely preserved along smooth training trajectories over any fixed horizon, and (iii) show how these constructions descend to moduli space $\Theta/G$ in the presence of internal symmetries. We then complement this theory with empirical geometric diagnostics that quantify, for concrete pretrained models, how close their finite-sample representations are to the injectivity boundary.

On the inversion side, our problem fits into the broader landscape of inverse problems \citep{sun2021inverse}, and more specifically language model inversion. Several works attempt to reconstruct prompts from outputs, logits, or log-probability sequences using trained inverters, typically achieving approximate or semantic reconstructions rather than exact inversion \citep{morris2023textembeddingsrevealalmost, nazir2025betterlanguagemodelinversion, morris2023languagemodelinversion}. These highlight privacy risks when probabilities or embeddings are exposed, but differ from our setting: they rely on supervised inverters and do not analyze the injectivity properties of internal hidden states in decoder-only LMs. A related line casts prompt construction as a discrete optimization problem (prompt search) guided by gradients or scores \citep{shin2020autopromptelicitingknowledgelanguage, wen2023hardpromptseasygradientbased, guo2025evopromptconnectingllmsevolutionary, sun2022bbtv2gradientfreefuturelarge, deng2022rlpromptoptimizingdiscretetext}. In contrast, we neither train an inverse model nor optimize prompts. Instead, we study when the forward map is generically injective and provide simple geometric quantities—margins and co-Lipschitz constants—that quantify, on a given finite prompt set, how close a model is to violating injectivity in practice.

\section{Analytic injectivity and persistence under training}

We work under the standard real-analytic (or piece-wise real-analytic) idealization of decoder-only Transformers: for a fixed context bound and vocabulary, the prompt set $\mathcal{S}$ is finite\footnote{Generic injectivity on infinite discrete spaces would require additional compactness or uniform convergence assumptions and is beyond our scope.}, and the last-token hidden state at layer $\ell$ defines a map
\begin{equation}
F^\ell_\theta : \mathcal{S} \to \mathbb{R}^d,\quad s \mapsto h^\ell_\theta(s),
\end{equation}
which is real-analytic in the parameters $\theta \in \Theta \subset \mathbb{R}^m$. In particular, our formal arguments apply most directly to architectures whose building blocks are real-analytic (e.g., GELU or SiLU activations together with LayerNorm and softmax). For piecewise-polynomial choices such as ReLU, an analogous treatment can be obtained in the framework of definable sets in an o-minimal structure, but we do not develop that extension here; throughout, we therefore treat ReLU-based stacks as lying slightly outside our formal scope, while still being covered at an informal, heuristic level by the same geometric picture. While each layer $\ell$ also produces hidden states for intermediate token positions, and one could define analogous injectivity and geometric diagnostics for those maps, we focus on last-token hidden states because in a causal decoder-only LM they are the representations used for next-token prediction and the only single positions that see the full prompt.\footnote{For each layer $\ell$ and token position $t$, consider the full hidden matrix $X^\ell_\theta(s) \in \mathbb{R}^{T \times d}$ (with $T$ the length of $s$), or position-specific maps $F^{\ell,t}_\theta(s) = X^{\ell, t}_\theta(s)$. One can run the same generic injectivity arguments for these objects, and construct corresponding empirical metrics. This might be useful for studying questions such as how quickly information about early tokens stabilizes with depth, or how context is compressed as one moves forward in the sequence.}

Following \citet{original_paper}, we study the \emph{collision locus} where distinct prompts share a representation. For each pair $s \neq \tilde s$ and each layer $\ell$, consider
\begin{equation}
H^\ell_{s,\tilde s}(\theta)\equiv F^\ell_\theta(s) - F^\ell_\theta(\tilde s)\,.
\end{equation}
This is a real-analytic in the parameters $\theta \in \Theta \subset \mathbb{R}^m$, and its zero set
\begin{equation}
Z^\ell_{s,\tilde s} = \{\theta\in\Theta: H^\ell_{s,\tilde s}=0\}\,,
\end{equation}
is either all of $\Theta$ (the degenerate case where $F^\ell_\theta(s)=F^\ell_\theta(\tilde s)$ for every $\theta$) or an analytic subset of codimension at least one. In the non-degenerate regime, each $Z^\ell_{s,\tilde s}$ therefore has Lebesgue measure zero. Taking the finite union over all unordered prompt pairs,
\begin{equation}
\Delta^\ell \equiv \bigcup_{s\neq\tilde s}Z^\ell_{s,\tilde s}\,,
\end{equation}
we obtain an analytic discriminant $\Delta^\ell \subset \Theta$ of Lebesgue measure zero. Its complement $\mathcal{U}^\ell\equiv\Theta\setminus\Delta^\ell$ is an open dense \emph{injective stratum} on which, for every $\theta \in \mathcal{U}^\ell$, the map $F^\ell_\theta$ is injective on $\mathcal{S}$. Thus, for each layer separately, either the model is nowhere injective on $\mathcal{S}$ (some pair collides for all $\theta$), or injectivity on $\mathcal{S}$ holds on an open dense subset of parameter space.

Beyond this static picture, we are interested in how injectivity behaves along training trajectories. Let $(\theta_t)_{t\geq 0}$ be the parameter sequence produced by an update rule 
\begin{equation}
\theta_{t+1} = T_t(\theta_t,\Xi_t)\,,
\end{equation}
where $\Xi_t$ encodes minibatch selection, stochastic noise and other randomness. Under these assumptions, the law of $\theta_t$ remains absolutely continuous for all $t$ up to any fixed horizon, obtained by iterating the pushforward and using that mixtures of absolutely continuous measures are absolutely continuous. Since each discriminant $\Delta^\ell$ has Lebesgue measure zero, it follows that with probability one
\begin{equation}
\theta_{t}\not\in\Delta^\ell \quad \mathrm{for\ all\ layers\ } \ell \mathrm{\ and\ all\ } t\leq T\,,
\end{equation}
so the parameter trajectory remains in the injective strata at all layers up to any fixed training time $T$. Smooth optimizers such as GD, SGD, or Adam applied to $\mathcal{C}^2$ losses satisfy this non-singularity condition almost everywhere in the usual real-valued idealization: their updates are of the form $\theta \mapsto \theta - \eta_t \nabla L_t(\theta)$, with Jacobian $I - \eta_t \nabla^2 L_t(\theta)$ having non-zero determinant outside a Lebesgue-null set where the Hessian has a specific spectrum. In this sense, generic injectivity is preserved along typical training runs in the continuous model of parameter space.

Assuming that the law of $\theta_0$ is absolutely continuous with respect to Lebesgue measure on $\Theta$, and that each $T_t(\cdot,\xi)$ is measurable and non-singular (preimages of Lebesgue-null sets are null), we show that the law of $\theta_t$ remains absolutely continuous for all $t$ up to any finite horizon. Since each discriminant $\Delta^\ell$ has Lebesgue measure zero, it follows that with probability one $\theta_t \in \mathcal{U}^\ell$ for all layers $\ell$ and all $t$ in this time window. Smooth optimizers such as GD/SGD/Adam on $\mathcal{C}^2$ losses satisfy these assumptions almost everywhere via their Jacobian, so generic injectivity is preserved along typical training runs in the continuous idealization. These assumptions deliberately abstract away from common non-smooth training heuristics such as hard parameter projections, explicit clipping of weights or gradients, and low-precision training updates. Such operations can destroy non-singularity at the level of Lebesgue measure and therefore fall outside the scope of our analysis. 

Finally, we extend the construction to the presence of internal symmetries. Many Transformer parameterizations admit a non-trivial group $G$ of reparameterizations (such as permutations and compatible rescalings of neurons or attention heads) that leave the realized function invariant. The group $G$ acts on $\Theta$, and the quotient $\Theta/G$ parametrizes functional equivalence classes. The collision discriminants and injective strata are invariant under this action and therefore descend to well-defined subsets of $\Theta/G$. Generic injectivity is thus naturally a property of the induced map on $\Theta/G$, rather than of any particular representative $\theta \in \Theta$. This aligns the analytic picture with the functional symmetries of modern architectures. Our result should thus be read as a statement about the idealized real-valued dynamics of smooth optimizers applied to non-singular updates; when aggressive clipping, projection steps, or discrete quantization are used inside the optimizer, the persistence guarantee may fail, and understanding injectivity in those regimes remains an open problem.

\subsection{Setup and collision strata}
\begin{defn}[Prompt set $\mathcal{S}$]
We define the {\it prompt set} $\mathcal{S}$ as the collection of all tokenized prompts (input sequences). For a fixed context bound $K$ and a finite vocabulary $V$, this set consists of all token sequences of length at most $K$,
\begin{equation}
\mathcal{S} = \bigcup_{k=1}^K V^k\,.
\end{equation}
In particular, $\mathcal{S}$ is finite, with $|\mathcal{S}| \equiv \sum_{k=1}^K |V|^k < \infty$.
\end{defn}

\begin{defn}[Parameter space $\Theta$]
Let $\Theta \subset \mathbb{R}^m$ denote the parameter space of models. We assume that $\Theta$ is a Borel (typically open) subset of $\mathbb{R}^m$, and equip it with the Lebesgue measure $\lambda$ induced from $\mathbb{R}^m$.
\end{defn}

\begin{defn}[Layer-wise forward maps $F^\ell_\theta$]
For $\theta \in \Theta$ and each layer index $\ell \in \{1,\dots,L\}$, the forward map $F^\ell_\theta$ sends a prompt $s \in \mathcal{S}$ to the last-token hidden state at layer $\ell$,
\begin{equation}
F^\ell_\theta : \mathcal{S} \to \mathbb{R}^d,\qquad s \mapsto h^\ell_\theta(s)\,.
\end{equation}
We assume that the architecture is such that, for each fixed $s \in \mathcal{S}$ and each layer $\ell$, the map $\theta \mapsto F^\ell_\theta(s)$ is real-analytic on $\Theta$. This holds for standard decoder-only Transformer blocks with real-analytic (or piece-wise real-analytic) nonlinearities (e.g.,\ GELU/SiLU/SwiGLU), LayerNorm with $\epsilon>0$, softmax-based causal attention, and standard positional encodings; see \citet{original_paper} for discussion. When we are only interested in the last layer, we denote $F^L_\theta$ simply by $F_\theta$.
\end{defn}

\begin{defn}[Collision sets $Z^\ell_{s,\tilde s}$]
For distinct prompts $s,\tilde s \in \mathcal{S}$, we define the {\it collision set} at layer $\ell$ by
\begin{equation}
Z^\ell_{s, \tilde s} \equiv \{\theta \in \Theta : F^\ell_\theta(s) = F^\ell_\theta(\tilde s)\}\,.
\end{equation}
This is exactly the set of parameter configurations for which the layer-$\ell$ map fails to be injective on the pair $\{s,\tilde s\}$.
\end{defn}

\begin{defn}[Discriminant $\Delta^\ell$]
The collision sets induce a {\it discriminant locus} at layer $\ell$,
\begin{equation}
\Delta^\ell = \bigcup_{s\neq \tilde s} Z^\ell_{s, \tilde s}\,.
\end{equation}
For any $\theta\in\Delta^\ell$, the map $F^\ell_\theta$ fails to be injective on $\mathcal S$, since at least one pair of distinct prompts shares the same last-token hidden state.
\end{defn}

\noindent{\bf Remark.}
Since $\mathcal{S}$ is finite and each $F^\ell_\theta(s)$ is real-analytic in $\theta$, the maps
\begin{equation}
H^\ell_{s,\tilde s}(\theta) \equiv F^\ell_\theta(s) - F^\ell_\theta(\tilde s)
\end{equation}
are real-analytic maps $\Theta \to \mathbb{R}^d$. For each fixed $(s,\tilde s)$, the common zero set
\begin{equation}
Z^\ell_{s,\tilde s} \;=\; \{\theta \in \Theta : H^\ell_{s,\tilde s}(\theta) = 0\}
\end{equation}
is therefore an analytic subset of $\Theta$. Either $H^\ell_{s,\tilde s}$ is identically zero, in which case $Z^\ell_{s,\tilde s} = \Theta$, or $H^\ell_{s,\tilde s}$ is not identically zero, in which case a standard fact from real-analytic geometry implies that $Z^\ell_{s,\tilde s}$ has (real) codimension at least one and hence Lebesgue measure zero. Since there are only finitely many unordered pairs $(s,\tilde s)$, the discriminant $\Delta^\ell$ is a finite union of such zero sets and thus also satisfies $\lambda(\Delta^\ell) = 0$. This motivates the study of the injective complement.

\begin{defn}[Injective stratum $\mathcal{U}^\ell$]
The injective stratum at layer $\ell$ is the complement of the discriminant,
\begin{equation}
\mathcal{U}^\ell \equiv \Theta \setminus \Delta^\ell\,.
\end{equation}
Each $Z^\ell_{s,\tilde s}$ is closed in $\Theta$ as the zero set of a continuous map, so $\Delta^\ell$ is closed as a finite union of closed sets and $\mathcal{U}^\ell$ is open. Moreover, whenever $\Delta^\ell \neq \Theta$ (i.e.\ we are not in the fully degenerate case where some pair collides for all $\theta$), $\Delta^\ell$ is a proper real-analytic subset of $\Theta$ and hence has empty interior; thus $\mathcal{U}^\ell$ is dense in $\Theta$. In this sense, for such layers, $F^\ell_\theta$ is \emph{generically} injective on $\mathcal{S}$ for $\theta \in \mathcal{U}^\ell$.
\end{defn}

\noindent{\bf Remark (persistence under training).}  
Suppose our update method is non-singular and preserves absolute continuity with respect to Lebesgue measure (in the sense formalized in Theorem~\ref{theorem:persistence} below). Since each $\Delta^\ell$ has Lebesgue measure zero, any parameter distribution that admits a density at initialization will, for any fixed finite training horizon, almost surely remain in the injective stratum $\mathcal{U}^\ell$ at every iteration (cf.\ Theorem~\ref{theorem:persistence}).

For any fixed $\theta \in \mathcal{U}^\ell$, each representation $x$ in the image $\mathrm{Im}(F^\ell_\theta) \subset \mathbb{R}^d$ has a unique preimage $s \in \mathcal{S}$. Any algorithm that recovers $s$ from $x = F^\ell_\theta(s)$ defines a (possibly partial) right inverse
\begin{equation}
\mathrm{inv}_\theta : \mathrm{Im}(F^\ell_\theta) \to \mathcal{S},\qquad \mathrm{inv}_\theta \circ F^\ell_\theta = \mathrm{id}_{\mathcal{S}}\,.
\end{equation}
In our setting we fix $\theta$ and invert the map $F^\ell_\theta : \mathcal{S} \to \mathbb{R}^d$, so the inverse is naturally a \emph{right} inverse. In \citet{original_paper} the authors instead consider an aggregate map that takes parameters as input and produces the tuple of outputs on a finite prompt set; there the corresponding inverse is a \emph{left} inverse with respect to that aggregate map. The two viewpoints are equivalent up to this change of perspective.

\subsection{Generic injectivity}
\begin{theorem}{\bf Layerwise generic injectivity}.
For each layer $\ell$, under the definitions above, either
\begin{itemize}
    \item the maps $F^\ell_\theta$ are non-injective on $\mathcal{S}$ for all $\theta \in \Theta$ (the fully degenerate case), or
    \item the discriminant $\Delta^\ell$ is a proper analytic subset of $\Theta$ with Lebesgue measure zero, $\lambda(\Delta^\ell) = 0$, and the injective stratum $\mathcal{U}^\ell$ is open and dense in $\Theta$, with $F^\ell_\theta : \mathcal{S} \to \mathbb{R}^d$ injective for all $\theta \in \mathcal{U}^\ell$.
\end{itemize}
In particular, if there exists at least one configuration $\theta$ for which $F^\ell_\theta$ is injective, then $F^\ell_\theta$ is generically injective on $\mathcal{S}$, i.e.\ injective for all $\theta$ in an open dense subset of $\Theta$.
\end{theorem}

\begin{proof}
Fix a layer $\ell$.

{\bf 1. Analytic zero sets.}  
For each unordered pair $(s,\tilde s)$ of distinct prompts, define
\begin{equation}
H^\ell_{s,\tilde s}(\theta) \equiv F^\ell_\theta(s) - F^\ell_\theta(\tilde s) \in \mathbb{R}^d\,.
\end{equation}
By assumption, $\theta \mapsto F^\ell_\theta(s)$ is real-analytic for each $s$, so each coordinate of $H^\ell_{s,\tilde s}$ is real-analytic. The common zero set
\begin{equation}
Z^\ell_{s,\tilde s} \;=\; \{\theta \in \Theta : H^\ell_{s,\tilde s}(\theta) = 0\}
\end{equation}
is therefore an analytic subset of $\Theta$.

{\bf 2. Excluding the degenerate case.} If $H^\ell_{s,\tilde s}$ is identically zero on $\Theta$, then $Z^\ell_{s,\tilde s} = \Theta$ and $F^\ell_\theta(s) = F^\ell_\theta(\tilde s)$ for all $\theta \in \Theta$. In that case $F^\ell_\theta$ is never injective on $\mathcal{S}$ for any $\theta$, and we are in the first case of the theorem.

Otherwise, $H^\ell_{s,\tilde s}$ is not identically zero. A standard fact about real-analytic maps then implies that $Z^\ell_{s,\tilde s}$ is an analytic subset of (real) codimension at least one and therefore has Lebesgue measure zero: $\lambda(Z^\ell_{s,\tilde s})=0$.

{\bf 3. Finite union of null sets.} Since $\mathcal{S}$ is finite, there are only finitely many unordered pairs $(s,\tilde s)$. It follows that
\begin{equation}
\Delta^\ell \equiv \bigcup_{s\neq\tilde s} Z^\ell_{s,\tilde s}
\end{equation}
is a finite union of measure-zero sets and hence itself has vanishing Lebesgue measure, $\lambda(\Delta^\ell)=0$.

{\bf 4. Open and dense injective stratum.} Each $Z^\ell_{s,\tilde s}$ is closed in $\Theta$ as the zero set of a continuous map, so their union $\Delta^\ell$ is closed and $\mathcal{U}^\ell = \Theta \setminus \Delta^\ell$ is open. If we are not in the degenerate case of step~2 (i.e.\ if there exists some $\theta$ for which $F^\ell_\theta$ is injective), then $\Delta^\ell \neq \Theta$ and is a proper analytic subset. Proper analytic subsets of $\mathbb{R}^m$ (or open subsets thereof) have empty interior, so $\Delta^\ell$ has empty interior and $\mathcal{U}^\ell$ is dense in $\Theta$.

{\bf 5. Injectivity on $\mathcal{S}$.} Finally, for any $\theta \in \mathcal{U}^\ell$, by definition there is no pair $s \neq \tilde s$ with $F^\ell_\theta(s) = F^\ell_\theta(\tilde s)$. Thus $F^\ell_\theta$ is injective on $\mathcal{S}$. 

This proves the dichotomy in the statement.
\end{proof}

\begin{cor}[Simultaneous generic injectivity across layers]

Suppose that for each layer $\ell \in \{1,\dots,L\}$, the map $F^\ell_\theta$ is injective on $\mathcal{S}$ for at least one $\theta \in \Theta$. Then for each $\ell$, the corresponding injective stratum $\mathcal{U}^\ell$ is non-empty, open, and dense in $\Theta$, and the intersection
\begin{equation}
\bigcap_{\ell=1}^L \mathcal{U}^\ell
\end{equation}
is dense (and non-empty) in $\Theta$. Any $\theta$ in this intersection induces injective maps on $\mathcal{S}$ for all layers simultaneously.
\end{cor}

\begin{proof}
By the theorem, for each $\ell$ the set $\mathcal{U}^\ell$ is open and dense and non-empty. Finite intersections of open dense sets are open and dense, so $\bigcap_{\ell=1}^L \mathcal{U}^\ell$ is open, dense, and non-empty in $\Theta$. For any $\theta$ in this intersection, $F^\ell_\theta$ is injective on $\mathcal{S}$ for all $\ell$ by the definition of $\mathcal{U}^\ell$.
\end{proof}

\subsection{Robust injectivity radius - safety margin to discriminant}
In this subsection, we isolate a simple notion of a \emph{robust injectivity radius} for maps on finite sets and relate it to a separation margin to the discriminant. This gives a clean way to quantify how much perturbation a representation map can tolerate before losing injectivity, and provides a natural lens on post-hoc distortions such as quantization. While straightforward for finite domains, we formalize this notion of a robust injectivity radius to connect the finite-set geometry to later empirical diagnostics.

\begin{defn}[Margin on a finite set]
Let $(S,d_X)$ be a finite metric space and let $F : S \to \mathbb{R}^d$ be any map. The (Euclidean) \emph{margin} of $F$ on $S$ is
\begin{equation}
m(F) \;\equiv\; \min_{s \neq \tilde s \in S} \big\|F(s) - F(\tilde s)\big\|_2\,.
\end{equation}
If $F$ is injective, then $m(F) > 0$.
\end{defn}

\begin{defn}[Robust injectivity radius]
Let $F : S \to \mathbb{R}^d$ be as above. For $r \ge 0$, we say that a map $\tilde F : S \to \mathbb{R}^d$ is an \emph{$r$-perturbation} of $F$ if
\begin{equation}
\label{eq:uniform-perturbation}
\|\tilde F - F\|_{\infty,S}
\;\equiv\;
\max_{s \in S} \big\|\tilde F(s) - F(s)\big\|_2
\;\le\; r.
\end{equation}
The \emph{robust injectivity radius} of $F$ on $S$ is
\begin{equation}
r_{\mathrm{inj}}(F)
\;\equiv\;
\sup\big\{ r \ge 0 : \text{every $r$-perturbation $\tilde F$ of $F$ is injective on $S$} \big\}.
\end{equation}
\end{defn}

Thus $r_{\mathrm{inj}}(F)$ is the largest perturbation radius such that \emph{all} perturbations of $F$ within that radius preserve injectivity on $S$. The next theorem identifies $r_{\mathrm{inj}}(F)$ exactly in terms of the margin.

\begin{theorem}[Robust injectivity radius equals half the margin]
\label{thm:robust-inj-radius}
Let $F : S \to \mathbb{R}^d$ be a map on a finite set $S$ with margin $m(F)$, and suppose $F$ is injective so that $m(F) > 0$. Then
\begin{equation}
r_{\mathrm{inj}}(F) \;=\; \frac{1}{2}\,m(F)\,.
\end{equation}
\end{theorem}

\begin{proof}
We prove the two inequalities $r_{\mathrm{inj}}(F) \ge m(F)/2$ and $r_{\mathrm{inj}}(F) \le m(F)/2$ separately.

\paragraph{Lower bound.}
Let $0 \le r < m(F)/2$ and let $\tilde F$ be any $r$-perturbation of $F$, i.e.\ $\|\tilde F - F\|_{\infty,S} \le r$. For any distinct $s,\tilde s \in S$, the triangle inequality gives
\begin{align}
\big\|\tilde F(s) - \tilde F(\tilde s)\big\|_2
&\ge \big\|F(s) - F(\tilde s)\big\|_2
   - \big\|\tilde F(s) - F(s)\big\|_2
   - \big\|\tilde F(\tilde s) - F(\tilde s)\big\|_2 \\
&\ge m(F) - 2r\,.
\end{align}
Since $r < m(F)/2$, we have $m(F) - 2r > 0$, hence $\tilde F(s) \neq \tilde F(\tilde s)$ for all $s \neq \tilde s$. Thus any $r$-perturbation with $r < m(F)/2$ is injective on $S$, and by definition of $r_{\mathrm{inj}}(F)$ this shows $r_{\mathrm{inj}}(F) \ge m(F)/2$.

\paragraph{Upper bound.}
Let $r > m(F)/2$. Choose a pair $(s_0,s_1) \in S \times S$ with $s_0 \neq s_1$ attaining the margin, i.e.
\begin{equation}
\big\|F(s_0) - F(s_1)\big\|_2 = m(F)\,.
\end{equation}
Define a perturbed map $\tilde F : S \to \mathbb{R}^d$ by
\begin{equation}
\tilde F(s) =
\begin{cases}
\frac{1}{2}\big(F(s_0) + F(s_1)\big), & \text{if } s \in \{s_0,s_1\},\\
F(s), & \text{otherwise.}
\end{cases}
\end{equation}
Then
\begin{equation}
\big\|\tilde F(s_0) - F(s_0)\big\|_2
= \big\|\tilde F(s_1) - F(s_1)\big\|_2
= \frac{1}{2}\,\big\|F(s_0) - F(s_1)\big\|_2
= \frac{1}{2}\,m(F),
\end{equation}
and $\tilde F(s) = F(s)$ for all $s \notin \{s_0,s_1\}$. Hence
\begin{equation}
\|\tilde F - F\|_{\infty,S}
= \max_{s\in S} \big\|\tilde F(s) - F(s)\big\|_2
= \frac{1}{2}\,m(F) < r\,.
\end{equation}
However, by construction $\tilde F(s_0) = \tilde F(s_1)$, so $\tilde F$ is not injective on $S$. Therefore, for any $r > m(F)/2$ there exists an $r$-perturbation of $F$ which is non-injective, and by definition of $r_{\mathrm{inj}}(F)$ this implies $r_{\mathrm{inj}}(F) \le m(F)/2$.

Combining the lower and upper bounds yields $r_{\mathrm{inj}}(F) = m(F)/2$, as claimed.
\end{proof}

In particular, for any layerwise last-token map $F^\ell_\theta : \mathcal{S} \to \mathbb{R}^d$ with margin $m^\ell(\theta)$, the robust injectivity radius on a finite prompt set $S \subset \mathcal{S}$ is exactly $m^\ell(\theta)/2$.

\paragraph{Application: uniform quantization.}
As a concrete illustration, consider the coordinate-wise uniform scalar quantizer. Let $Q_b : \mathbb{R}^d \to \mathbb{R}^d$ denote a uniform quantizer with step size $\delta_b > 0$, applied coordinate-wise:
\begin{equation}
(Q_b(x))_k = \delta_b \cdot \mathrm{round}\!\Big(\frac{x_k}{\delta_b}\Big),\qquad k = 1,\dots,d.
\end{equation}
Each coordinate moves by at most $\delta_b/2$ under quantization, so for any $x \in \mathbb{R}^d$ we have
\begin{equation}
\big\|Q_b(x) - x\big\|_2 \;\le\; \sqrt{d}\,\frac{\delta_b}{2}\,.
\end{equation}
Given a representation map $F : S \to \mathbb{R}^d$, we define its quantized version on $S$ by
\begin{equation}
F_b(s) \;\equiv\; Q_b(F(s)).
\end{equation}
Then
\begin{equation}
\|F_b - F\|_{\infty,S}
= \max_{s\in S} \big\|Q_b(F(s)) - F(s)\big\|_2
\;\le\; \sqrt{d}\,\frac{\delta_b}{2}\,.
\end{equation}
Combining this with Theorem~\ref{thm:robust-inj-radius} yields the following simple sufficient condition for quantization to preserve injectivity.

\begin{cor}[Sufficient condition for injectivity under quantization]
\label{cor:quantization-injectivity}
Let $F : S \to \mathbb{R}^d$ be injective with margin $m(F) > 0$, and let $F_b = Q_b \circ F$ be its quantized version with step size $\delta_b$. If
\begin{equation}
\sqrt{d}\,\frac{\delta_b}{2} < \frac{1}{2}\,m(F)
\quad\Longleftrightarrow\quad
\delta_b < \frac{m(F)}{\sqrt{d}}\,,
\end{equation}
then $F_b$ is injective on $S$.
\end{cor}

In particular, for each layer $\ell$, prompt set $S$ and parameter $\theta$, if we denote by $m^\ell(\theta)$ the separation margin of $F^\ell_\theta$ on $S$, then any uniform quantization with step size
\begin{equation}
\delta_b < \frac{m^\ell(\theta)}{\sqrt{d}}
\end{equation}
is guaranteed not to introduce collisions on $S$. Equivalently, the onset of quantization-induced collisions occurs at a layerwise, parameter-dependent critical bitwidth $b^\ell_{\mathrm{crit}}(\theta)$ determined by the layerwise margin and hidden dimension. Later, in the empirical section, we will estimate $m^\ell(\theta)$ from finite samples, which provides a practical handle on how close a given configuration is to this critical regime.

\subsection{Persistence under training}
\begin{theorem}{\bf Persistence of generic injectivity under smooth non-singular updates (idealized continuous setting)}.
\label{theorem:persistence}

\begin{enumerate}[label=(\roman*)]
\item Let $\Delta = \bigcup_{\ell=1}^L \Delta^\ell$ and $\mathcal{U} = \Theta \setminus \Delta$ denote the union of layerwise discriminants and its complement. Consider a (possibly random) sequence of parameter updates
\begin{equation}
\theta_{t+1} = T_t(\theta_t, \Xi_t)\,,\qquad t=0,1,\dots,
\end{equation}
where $(\Xi_t)_{t\geq 0}$ is a sequence of exogenous random variables (independent of $\theta_t$ conditional on the past) encoding non-parametric update information such as minibatch selection, injected noise, or dropout masks. We assume that each $\Xi_t$ takes values in some measurable space $(\mathsf{X}, \mathcal{F})$; no additional structure on $\mathsf{X}$ is required, and this formalism simply allows us to treat all sources of randomness uniformly.

For each $t$ and each $\xi \in \mathsf{X}$, we write
\begin{equation}
T_{t,\xi} : \Theta \to \Theta,\qquad \theta \mapsto T_t(\theta,\xi)
\end{equation}
for the update map with the randomness held fixed, and assume that $T_{t,\xi}$ is measurable in $\theta$.

\item Let $\theta_0 \in \Theta$ be a random initialization whose distribution $\mu_0$ is absolutely continuous with respect to Lebesgue measure $\lambda$ on $\Theta$, i.e.\ $\mu_0 \ll \lambda$ (so $\lambda(A)=0 \Rightarrow \mu_0(A)=0$ for all measurable $A \subseteq \Theta$). Assume that, for each step $t$ and for all $\xi$ in a set of full measure with respect to the law of $\Xi_t$, the map $T_{t,\xi}$ is non-singular with respect to Lebesgue, in the sense that
\begin{equation}
\lambda(A)=0 \;\Longrightarrow\; \lambda\bigl(T^{-1}_{t,\xi}(A)\bigr)=0
\end{equation}
for all measurable sets $A \subseteq \Theta$.
\end{enumerate}

\noindent Then for any fixed finite step horizon $T$, the probability
\begin{equation}
\mathbb{P}(\theta_t\in\mathcal{U}\,\,\forall\,t\leq T)=1\,.
\end{equation}
Equivalently, if $\mu_t$ denotes the distribution of $\theta_t$, then $\mu_t(\Delta) = 0$ for all $t \leq T$.

This result holds in the standard continuous parameter idealization of training. In practical floating-point implementations, strict non-singularity is not guaranteed, but the result characterizes the almost-sure behavior in the continuous limit.
\end{theorem}

\begin{proof}

{\bf 1. Conditional pushforward.}  
Fix $t \geq 0$. For a fixed realization $\xi \in \mathsf{X}$ of $\Xi_t$, the conditional update is $\theta_{t+1} = T_{t,\xi}(\theta_t)$, so the conditional distribution of $\theta_{t+1}$ given $\Xi_t = \xi$ is the pushforward
\begin{equation}
\mu^{\xi}_{t+1} = (T_{t,\xi})_{\#} \mu_t\,.
\end{equation}

{\bf 2. Preservation of absolute continuity under non-singular maps.}  
Since $T_{t,\xi}$ is non-singular and $\mu_t \ll \lambda$, it follows that $\mu^\xi_{t+1} \ll \lambda$ for all $\xi$ in the full-measure subset where $T_{t,\xi}$ is non-singular. Indeed, if $\lambda(A)=0$ then $\lambda(T^{-1}_{t,\xi}(A))=0$ and hence
\begin{equation}
\mu^\xi_{t+1}(A) = \mu_t\bigl(T^{-1}_{t,\xi}(A)\bigr) = 0.
\end{equation}

{\bf 3. Mixtures of absolutely continuous measures.}
Let $\nu_t$ denote the law of $\Xi_t$. By conditioning on $\Xi_t$ and using the law of total probability, the unconditional distribution of $\theta_{t+1}$ is the mixture
\begin{equation}
\mu_{t+1} \;=\; \int_{\mathsf{X}} \mu^\xi_{t+1}\, \mathrm{d}\nu_t(\xi)\,.
\end{equation}
In particular, for any measurable $A \subseteq \Theta$,
\begin{equation}
\mu_{t+1}(A) \;=\; \int_{\mathsf{X}} \mu^\xi_{t+1}(A)\, \mathrm{d}\nu_t(\xi).
\end{equation}
Since each $\mu^\xi_{t+1} \ll \lambda$, we have $\lambda(A)=0 \Rightarrow \mu^\xi_{t+1}(A)=0$ for all $\xi$, and hence $\mu_{t+1}(A)=0$. Thus $\mu_{t+1} \ll \lambda$.

\textbf{4. Induction.}  
By induction on $t$, starting from $\mu_0 \ll \lambda$, we obtain $\mu_t \ll \lambda$ for all $t \leq T$.

{\bf 5. Avoiding the discriminant.}  
Since $\Delta$ has vanishing Lebesgue measure, $\lambda(\Delta)=0$, absolute continuity implies that $\mu_t(\Delta) = 0$ for all $t \leq T$. Equivalently, $\mathbb{P}(\theta_t \in \mathcal{U}) = 1$ for each $t \leq T$. Taking the union over the finitely many steps $t \in \{0,\dots,T\}$ shows that
\begin{equation}
\mathbb{P}\bigl(\theta_t \in \mathcal{U} \text{ for all } t \leq T\bigr) = 1.
\end{equation}

\end{proof}

\begin{cor}{\bf Injective stratum permanence}
Under the conditions of Theorem~\ref{theorem:persistence}, if we view $(\theta_t)_{t=0}^T$ as random variables taking values in $\Theta$, then with probability~1 the model never leaves the injective stratum up to time $T$, i.e.\ $\theta_t \in \mathcal{U}$ for all $t \leq T$ almost surely.
\end{cor}

\begin{proof}
Immediate from Theorem~\ref{theorem:persistence} and the definition of $\mathcal{U}$.
\end{proof}

\begin{lem}[Smooth GD is non-singular a.e.]
Let $L : \Theta \to \mathbb{R}$ be $\mathcal{C}^2$ (or real-analytic). Fix $\eta > 0$ and define the gradient descent update
\begin{equation}
T(\theta) \;=\; \theta - \eta \nabla L(\theta)\,.
\end{equation}
If the Jacobian matrix $J(\theta) = I - \eta \nabla^2 L(\theta)$ satisfies $\det J(\theta) \neq 0$ for $\lambda$-almost every $\theta \in \Theta$, then $T$ is non-singular with respect to Lebesgue measure.
\end{lem}

\begin{proof}
The map $T$ is $\mathcal{C}^1$ by assumption. Wherever $\det J(\theta) \neq 0$, the inverse function theorem implies that $T$ is a local $\mathcal{C}^1$ diffeomorphism near $\theta$, and the change-of-variables formula for integrals shows that preimages of Lebesgue-null sets have Lebesgue measure zero in that region. By hypothesis, the set $\{\theta : \det J(\theta) = 0\}$ has Lebesgue measure zero; in particular, it does not affect the ``almost everywhere'' non-singularity property. Thus $T$ is non-singular with respect to Lebesgue measure.\footnote{If $L$ is real-analytic and $\det J$ is not identically zero, then the zero set $\{\theta : \det J(\theta)=0\}$ is a proper analytic subset and hence has Lebesgue measure zero. In that case the assumption in the lemma follows from analyticity.}
\end{proof}

\noindent{\bf Remark.}
The lemma shows that, under mild conditions on $\nabla^2 L$, gradient descent updates fit the non-singularity framework of Theorem~\ref{theorem:persistence}. In practice, optimizers such as GD/SGD/Adam/AdamW are implemented in finite-precision arithmetic, where the parameter space is effectively a finite grid and every update is singular in a strict measure-theoretic sense. Our analysis, however, is conducted in the standard continuous idealization in which parameters and optimizer states live in $\mathbb{R}^m$ and updates are given by smooth maps. In this setting, the non-singularity and absolute-continuity assumptions of Theorem~\ref{theorem:persistence} are satisfied for essentially any ``regular'' optimizer that does not introduce hard projections, clipping, or quantization of the parameters.

Formally, one can treat the optimizer state as part of an extended variable $(\theta,z)$, where $z$ collects momentum, moments, and other internal statistics, and write the update as
\begin{equation}
(\theta_{t+1}, z_{t+1}) \;=\; T_t(\theta_t, z_t, \Xi_t)\,.
\end{equation}
For fixed $\Xi_t$ (minibatch, randomness), the map
\begin{equation}
(\theta,z) \mapsto T_t(\theta,z,\Xi_t)
\end{equation}
is typically $\mathcal{C}^1$, and its Jacobian with respect to $(\theta,z)$ has non-vanishing determinant for Lebesgue-almost every $(\theta,z)$. This implies that $T_t$ is non-singular with respect to Lebesgue measure a.e.\ on the extended space, and therefore preserves absolute continuity of the joint distribution of $(\theta_t,z_t)$, so that the conclusions of Theorem~\ref{theorem:persistence} apply.

\subsection{Symmetries}

\begin{theorem}[Injectivity modulo internal symmetries]
\label{thm:symmetry}
Let $G$ be a measurable group (typically finite or locally compact) acting smoothly or continuously on $\Theta$ by measurable maps
\begin{equation}
G\times\Theta \to \Theta,\qquad (g,\theta)\mapsto g\cdot\theta,
\end{equation}
such that the forward maps are $G$-invariant in the sense that
\begin{equation}
F^\ell_{g\cdot\theta}(s) \;=\; F^\ell_\theta(s)\quad\text{for all }g\in G,\;\theta\in\Theta,\;s\in\mathcal S,\;\ell\in\{1,\dots,L\}.
\label{eq:G_invariance}
\end{equation}
Then:
\begin{enumerate}
    \item The discriminants $\Delta^\ell$ and injective strata $\mathcal U^\ell$ are $G$-invariant subsets of $\Theta$.
    \item The quotient map $\pi:\Theta\to\mathcal M=\Theta/G$ induces well-defined subsets
    \begin{equation}
    \overline{\Delta}^\ell = \pi(\Delta^\ell)\subset\mathcal M,\qquad \overline{\mathcal U}^\ell = \mathcal M\setminus \overline{\Delta}^\ell,
    \end{equation}
    which we call the discriminant and injective stratum in moduli space.
    \item Generic injectivity is a property of equivalence classes: if one representative $\theta$ of a class $[\theta]\in\mathcal M$ lies in $\mathcal U^\ell$ (so that $F^\ell_\theta$ is injective on $\mathcal S$), then every representative $\theta'\in[\theta]$ also lies in $\mathcal U^\ell$ and induces the same injective map on $\mathcal S$.
\end{enumerate}
In particular, under the assumptions of Theorem~\ref{theorem:persistence}, the iterative dynamics of a symmetry-respecting optimizer (one satisfying $T_t(g\cdot\theta,\xi) = g\cdot T_t(\theta,\xi)$ for all $g \in G$) induces a trajectory in moduli space that almost surely remains in the injective stratum $\bigcap_\ell \overline{\mathcal{U}}^\ell$ up to any fixed finite horizon. Generic injectivity is thus naturally a statement about the realized functions $F^\ell_\theta$ (equivalence classes in $\mathcal{M}$) rather than individual parameter vectors.
\end{theorem}

\begin{proof}[Proof sketch]
The $G$-invariance \eqref{eq:G_invariance} implies that for any $g \in G$ and any collision $(\theta,s,\tilde s)$ with $F^\ell_\theta(s) = F^\ell_\theta(\tilde s)$, we also have
\begin{equation}
F^\ell_{g\cdot\theta}(s) \;=\; F^\ell_\theta(s) \;=\; F^\ell_\theta(\tilde s) \;=\; F^\ell_{g\cdot\theta}(\tilde s),
\end{equation}
so $g\cdot\theta \in Z^\ell_{s,\tilde s}$. Hence each $Z^\ell_{s,\tilde s}$, and therefore $\Delta^\ell$ as their union, is $G$-invariant. Its complement $\mathcal{U}^\ell = \Theta \setminus \Delta^\ell$ is also $G$-invariant. The quotient map $\pi : \Theta \to \mathcal{M} = \Theta/G$ identifies points on the same $G$-orbit, so images of $G$-invariant sets are well-defined subsets of $\mathcal{M}$, giving $\overline{\Delta}^\ell$ and $\overline{\mathcal{U}}^\ell$.

If $\theta \in \mathcal{U}^\ell$, then $F^\ell_\theta$ is injective on $\mathcal{S}$ by definition. For any $g \in G$, condition \eqref{eq:G_invariance} implies $F^\ell_{g\cdot\theta} = F^\ell_\theta$ as functions on $\mathcal{S}$, so $F^\ell_{g\cdot\theta}$ is also injective. Thus every representative of the class $[\theta]$ lies in $\mathcal{U}^\ell$, and the induced map $F^\ell_{[\theta]} := F^\ell_\theta$ is well defined on $\mathcal{S}$ independently of the chosen representative. The persistence statement in moduli space then follows from Theorem~\ref{theorem:persistence} applied to $\Delta = \bigcup_\ell \Delta^\ell$ together with the assumed $G$-invariance of the dynamics.
\end{proof}

A full measure-theoretic treatment would require additional regularity assumptions on $G$ and the quotient $\Theta / G$ (e.g., local compactness or properness of the action); our argument implicitly assumes such regularity.

\section{Empirical study}
\label{sec:empirical}

In this section, we connect the analytic picture to concrete models. We fix a mixed prompt distribution and empirically probe the geometry of last-token representations in several pretrained decoder-only Transformers, focusing on (i) layerwise separation and co-Lipschitz behavior, (ii) dependence on sequence length, (iii) model scaling, and (iv) robustness to non-analytic perturbations such as post-hoc quantization of hidden states.\newline

All experiments were run on AMD Instinct\texttrademark\ MI300X GPU, with PyTorch in bf16.\footnote{We used the pre-built ROCm Docker image \href{https://hub.docker.com/layers/rocm/pytorch-training/v25.8/images/sha256-5082ae01d73fec6972b0d84e5dad78c0926820dcf3c19f301d6c8eb892e573c5}{\texttt{rocm/pytorch-training:v25.8}}.} The hardware and software stack affect only runtime and capacity; the qualitative geometric phenomenon we report are independent of the specific accelerator.

\subsection{Data and prompt construction}
\label{subsec:data}

We construct the experimental prompt set $\mathcal S_{\mathrm{exp}}\subset\mathcal S$ as follows.
We first build a mixed corpus of raw texts by sampling from standard sources (IMDB reviews, C4 subset), removing very short strings and de-duplicating at the string level.\footnote{While these sources do not match the full proprietary pretraining mixtures used for LLaMA-3 or Qwen, they do provide a diverse mix of natural English text with heterogeneous styles and lengths. Our goal is not to reproduce the exact pretraining distribution, but to probe the finite-set geometry of representations on a reasonably realistic prompt mixture; all injectivity and robustness statements below should be interpreted with this distributional restriction in mind.} This yields a fixed list of $N_{\mathrm{raw}}=10^5$ texts, which we reuse across all models and experiments. While $N_{\mathrm{raw}}=10^5$ provides a large and diverse enough sample for our purposes, it remains a tiny subset of the full combinatorial prompt space. Thus the absence of collisions empirically supports, but does not prove, injectivity for the full set $\mathcal{S}$.

Given a tokenizer associated with a particular model, we obtain tokenized prompts by applying the tokenizer with truncation at a global maximum length $K_{\max}$ (e.g.,\ $K_{\max}=1024$ tokens). For the fixed-length experiments we additionally form, for each $K \leq K_{\max}$, a prompt set
\begin{equation}
\mathcal{S}_{K} \subset \mathcal{S},
\end{equation}
by taking $K$-token prefixes of all tokenized texts of length at least $K$, truncating to length $K$, and then de-duplicating in token space. This produces a finite set of distinct token sequences of fixed length $K$ with no padding.

For a given model and choice of $K$, we denote the resulting experimental prompt set by
\begin{equation}
\mathcal{S}_{\mathrm{exp}}(K) = \{s_1,\dots,s_{N_K}\}\subset\mathcal{S}_K,
\end{equation}
where $N_K$ is the number of distinct $K$-token sequences after de-duplication.
We represent each $s_i$ by its token sequence
\begin{equation}
s_i \equiv (t_{i,1},\dots,t_{i,K}),\qquad t_{i,t} \in V,
\end{equation}
so that discrete input distances such as Hamming distance are well defined.

\subsection{Models and hidden-state extraction}
\label{subsec:models}

We consider pretrained decoder-only language models from the LLaMA-3 and Qwen families, spanning a range of sizes (from 0.5B to 8B parameters). For each model we denote by $L$ the number of layers and by $d$ the hidden dimension. Given a prompt $s_i \in \mathcal{S}_{\mathrm{exp}}(K)$, we write
\begin{equation}
x^\ell_i \equiv h^\ell_\theta(s_i) \in \mathbb{R}^d
\end{equation}
for the last-token hidden state at layer $\ell \in \{1,\dots,L\}$. Collecting these into matrices
\begin{equation}
X^\ell \in \mathbb{R}^{N_K \times d},\qquad X^\ell_{i,:} = x^\ell_i,
\end{equation}
allows us to view each layer as a finite point cloud in $\mathbb{R}^d$.

All experiments are implemented in PyTorch using the HuggingFace \texttt{transformers} library. Hidden states are extracted by running the models in evaluation mode with \texttt{output\_hidden\_states=True} and gathering, for each layer, the hidden vector at the last prompt position.

\subsection{Geometric diagnostics: margins, co-Lipschitz constants, and normalization}
\label{subsec:metrics}

To connect to the theoretical notions of injectivity and co-Lipschitz behavior, we define empirical diagnostics on $\mathcal S_{\mathrm{exp}}(K)$.

\paragraph{Nearest-neighbor separation (margin).}
For a fixed layer $\ell$ and model parameters $\theta$, the ideal separation margin is
\begin{equation}
\mathrm{margin}_\ell(\theta) 
\;\equiv\;
\min_{s\neq\tilde s} \bigl\|F^\ell_\theta(s) - F^\ell_\theta(\tilde s)\bigr\|_2.
\end{equation}
On the finite set $\mathcal S_{\mathrm{exp}}(K)$ we approximate this by nearest-neighbor statistics.
For each $i\in\{1,\dots,N_K\}$ we define the nearest-neighbor distance
\begin{equation}
d^\ell_i \;=\; \min_{j\neq i} \,\bigl\|x^\ell_i - x^\ell_j\bigr\|_2,
\end{equation}
computed exactly from $X^\ell$.\footnote{For the prompt set sizes we consider (typically $N_K \lesssim 10^5$), an exact $O(N_K^2)$ computation per layer is feasible with batching; approximate nearest neighbor methods could also be used.}
Rather than taking a strict minimum, which is sensitive to outliers and numerical noise, we define a worst-percentile empirical margin
\begin{equation}
\widehat{\mathrm{margin}}_\ell^{(q)} 
\;=\; \mathrm{quantile}_q\bigl(\{d^\ell_i\}_{i=1}^{N_K}\bigr),
\end{equation}
where $q$ is a small percentile (e.g.,\ $q=1\%$). This captures the typical separation among the hardest-to-separate prompts in $\mathcal S_{\mathrm{exp}}(K)$.

\paragraph{Co-Lipschitz constant.}
To quantify sensitivity of representations to changes in the prompt, we equip the prompt set with the Hamming distance
\begin{equation}
d_{\mathrm{in}}(s_i,s_j) \;=\; \#\{t : t_{i,t} \neq t_{j,t}\},
\end{equation}
i.e.\ the number of positions at which two $K$-token sequences differ. The theoretical co-Lipschitz (lower Lipschitz) constant at layer $\ell$ is
\begin{equation}
\alpha_\ell(\theta)
\;\equiv\;
\inf_{s\neq\tilde s} \frac{\bigl\|F^\ell_\theta(s) - F^\ell_\theta(\tilde s)\bigr\|_2}{d_{\mathrm{in}}(s,\tilde s)}.
\end{equation}
On the finite set $\mathcal S_{\mathrm{exp}}(K)$ we estimate this via a random subset of pairs. For each layer $\ell$ we sample a large set $\mathcal P$ of index pairs $(i,j)$ with $i\neq j$, discard pairs with $d_{\mathrm{in}}(s_i,s_j)<d_{\min}$ (e.g.,\ $d_{\min}=1$ or $2$), and compute the ratios
\begin{equation}
r^\ell_{i,j} \;=\; \frac{\|x^\ell_i - x^\ell_j\|_2}{d_{\mathrm{in}}(s_i,s_j)}.
\end{equation}
We then define the empirical worst-percentile co-Lipschitz constant
\begin{equation}
\widehat{\alpha}_\ell^{(q)} \;=\; \mathrm{quantile}_q\bigl(\{r^\ell_{i,j} : (i,j)\in\mathcal P\bigr),
\end{equation}
with $q$ again small (e.g.,\ $1\%$). Intuitively, $\widehat{\alpha}_\ell^{(q)}$ measures the typical amount by which the representation moves, per unit of Hamming distance, in the most contractive cases.

\paragraph{Norm normalization and scale-invariant metrics.}
Absolute margins and co-Lipschitz estimates can be strongly influenced by global scaling of the hidden states (e.g.,\ growth of $\|x^\ell_i\|_2$ with depth). To obtain scale-invariant diagnostics we also track the mean norm
\begin{equation}
\overline{\rho}_\ell \;=\; \frac{1}{N_K}\sum_{i=1}^{N_K}\|x^\ell_i\|_2
\end{equation}
and form normalized quantities
\begin{equation}
\widetilde{\mathrm{margin}}_\ell^{(q)} 
= \frac{\widehat{\mathrm{margin}}_\ell^{(q)}}{\overline{\rho}_\ell},
\qquad
\widetilde{\alpha}_\ell^{(q)} 
= \frac{\widehat{\alpha}_\ell^{(q)}}{\overline{\rho}_\ell}.
\end{equation}
These are invariant under global rescaling $x^\ell_i\mapsto c\,x^\ell_i$ and can be interpreted as margins and co-Lipschitz constants on the unit sphere; they highlight intrinsic geometric conditioning rather than overall energy.

\paragraph{Defaults.} Unless noted, we use $q=1$\%, $d_{\mathrm{min}}=1$ and $|\mathcal{P}|=5\times10^4$ sampled pairs per layer.

\paragraph{Uncertainty.} We estimate uncertainty via nonparametric bootstrap over prompts (for margins) and over pair samples $\mathcal{P}$ (for co-Lipschitz): we resample with replacement 200 times and report 2.5–97.5\% percentile intervals for $\widetilde{\mathrm{margin}}_\ell^{(q)}$ and $\widetilde{\alpha}^{(q)}_\ell$. These are plotted as shaded regions in relevant figures.

\paragraph{Collision counts.}
We directly test injectivity on $\mathcal{S}_{\mathrm{exp}}(K)$ by checking for exact collisions. For each layer $\ell$ we hash the floating-point vectors $\{x^\ell_i\}$ (or their quantized versions in the quantization experiments) and count distinct indices $i \neq j$ with $x^\ell_i = x^\ell_j$ to numerical precision (we treat two vectors as equal if they are bitwise identical in memory). After de-duplicating token sequences, any such collision is a genuine violation of injectivity on $\mathcal{S}_{\mathrm{exp}}(K)$ for the representation under consideration (full-precision or quantized).

\paragraph{Near-collision analysis.} In addition to exact (bitwise) collisions, we compute near-collision rates by declaring a pair $(i,j)$ a near collision if $\|x^\ell_i-x^\ell_j\|_2 \leq \varepsilon$ for $\varepsilon\in\{10^{-6}, 10^{-4}, 10^{-2}\}$ (full precision) and analogously for quantized states. These rates complement exact collision counts by revealing “practically non-separable” pairs at finite tolerance. Unless otherwise stated, we report the 
1\% worst-percentile nearest-neighbor distances and the fraction of pairs within each $\varepsilon$ at each layer. We visualize these layerwise near-collision fractions (for FP and 4-bit quantization) in Figure~\ref{fig:llama8b-eps-sweeps}.

For all models and prompt sets we tested, we observed \emph{no} collisions in full precision and under 8-bit post-hoc activation quantization. Under more aggressive 4-bit activation quantization, however, we do observe a modest number of collisions in the deepest layers (see Section~\ref{subsec:quantization}), reflecting the fact that the non-analytic quantizer $Q_4$ itself is non-injective, even though the underlying continuous map $F^\ell_\theta$ remains injective in full precision. All empirical metrics are evaluated on finite sampled prompt sets, so they approximate, rather than exactly characterize, the theoretical quantities $m(F)$ and $\alpha(F)$.

\subsection{Experimental setup}
\label{subsec:protocols}

We now describe the main empirical protocols used in the paper. Each is instantiated for several models and prompt sets, but the underlying procedures are shared.

\subsubsection{Layerwise geometry on a fixed prompt set}

\begin{figure*}[ht!]
    \centering

    \begin{subfigure}[b]{0.32\textwidth}
        \centering
        \includegraphics[width=\textwidth]{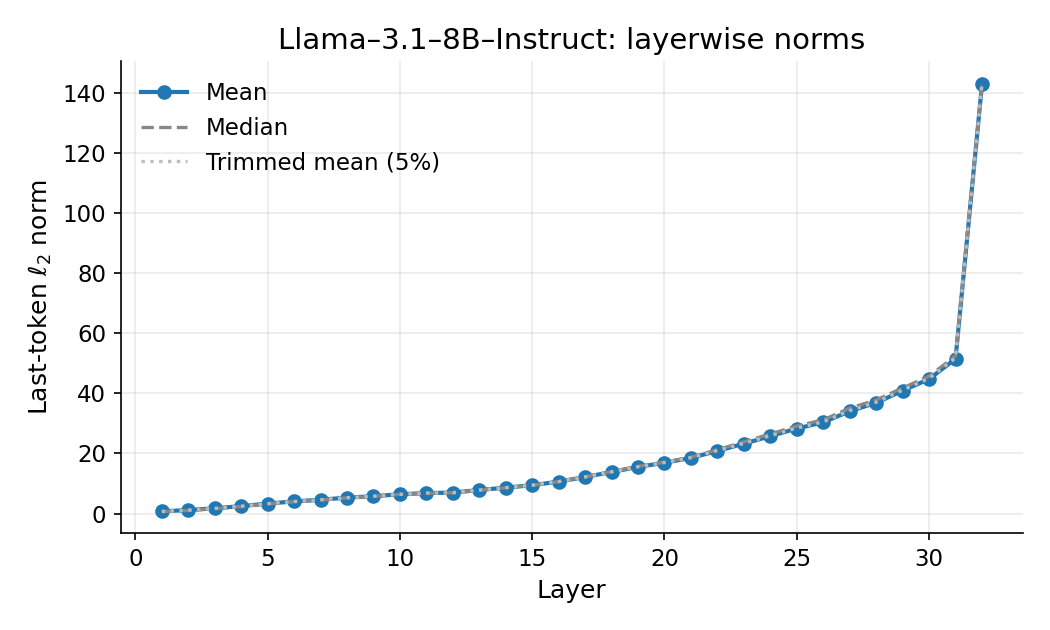}
        \subcaption{Mean $\ell_2$ norms}
        \label{fig:llama8b-layerwise-norms}
    \end{subfigure}
    \hfill
    \begin{subfigure}[b]{0.32\textwidth}
        \centering
        \includegraphics[width=\textwidth]{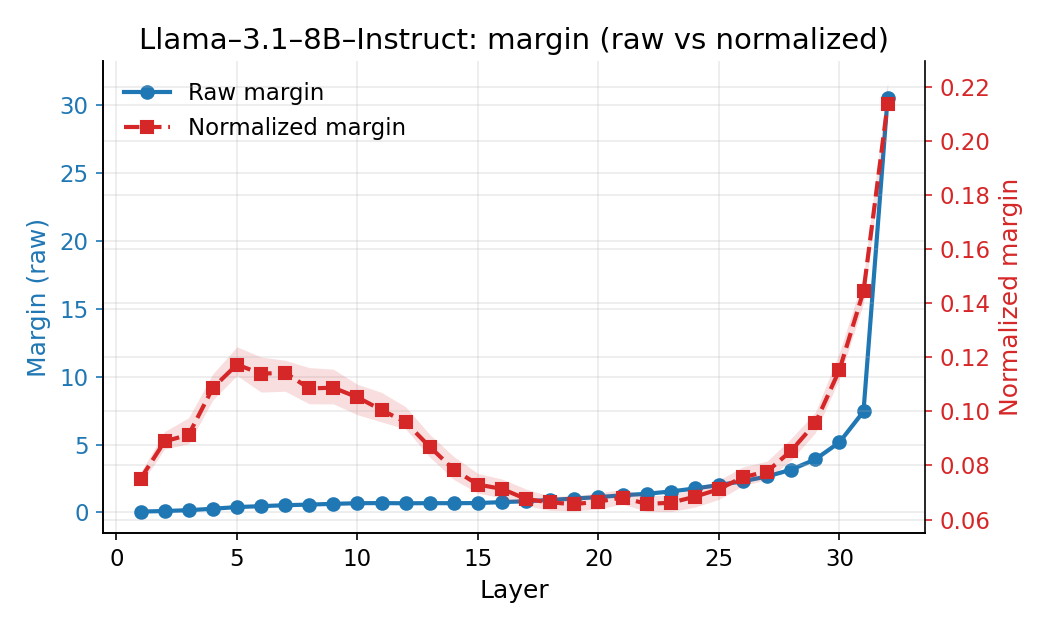}
        \subcaption{Separation margins.}
        \label{fig:llama8b-layerwise-margins}
    \end{subfigure}
    \hfill
    \begin{subfigure}[b]{0.32\textwidth}
        \centering
        \includegraphics[width=\textwidth]{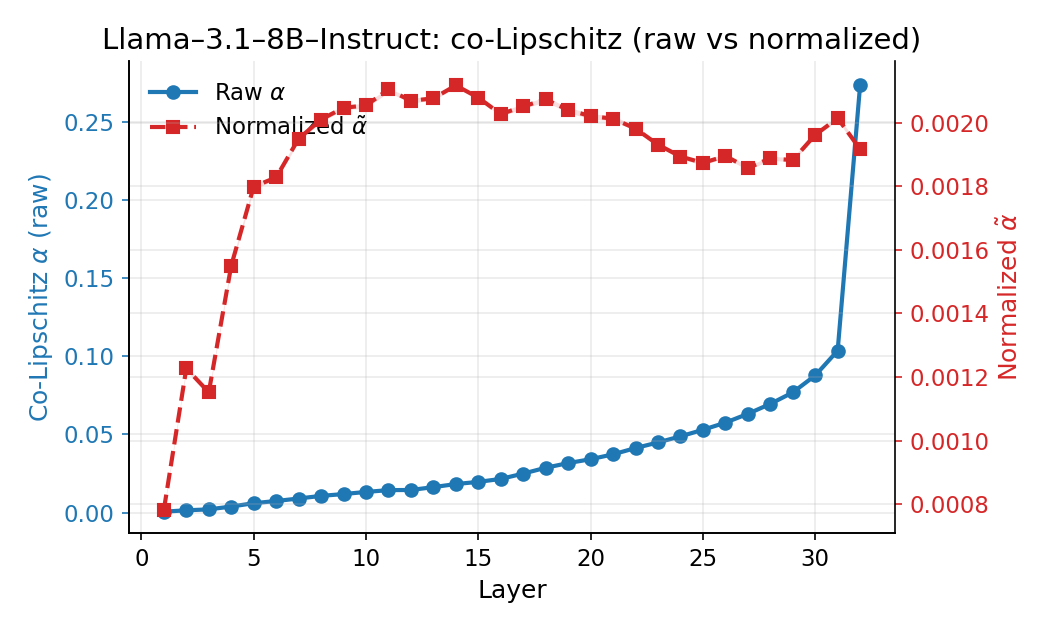}
        \subcaption{Co-Lipschitz estimates}
        \label{fig:llama8b-layerwise-alphas}
    \end{subfigure}

    \caption{Layerwise geometry of last-token representations for \texttt{Llama-3.1-8B-Instruct}. Margin and co-Lipschitz values are always plotted for $q = 1$\% worst-percentile. The raw values of margins and co-Lipschitz estimates are strongly influenced by the growing $\ell_2$ norms through the layers.}
    \label{fig:llama8b-layerwise-geometry}
\end{figure*}

For each model, we fix a base context length $K_{\mathrm{base}}$ (e.g.,\ $K_{\mathrm{base}}=512$) and construct $\mathcal{S}_{\mathrm{exp}}(K_{\mathrm{base}})$ as in Section~\ref{subsec:data}. We then extract last-token states $\{x^\ell_i\}$ at all layers $\ell \in \{1,\dots,L\}$ and compute, for each layer:
\begin{itemize}
  \item the mean norm $\overline{\rho}_\ell$;
  \item the worst-percentile margin $\widehat{\mathrm{margin}}_\ell^{(q)}$ and its normalized version $\widetilde{\mathrm{margin}}_\ell^{(q)}$;
  \item the worst-percentile co-Lipschitz constant $\widehat{\alpha}_\ell^{(q)}$ and its normalized version $\widetilde{\alpha}_\ell^{(q)}$;
  \item the number of exact representation collisions on $\mathcal S_{\mathrm{exp}}(K_{\mathrm{base}})$.
\end{itemize}
In Figure~\ref{fig:llama8b-layerwise-geometry} we visualize these quantities as layerwise curves $\ell\mapsto(\overline{\rho}_\ell,\widehat{\mathrm{margin}}_\ell^{(q)},\widetilde{\mathrm{margin}}_\ell^{(q)},\widehat{\alpha}_\ell^{(q)},\widetilde{\alpha}_\ell^{(q)})$, in order to illustrate how separation and co-Lipschitz behavior evolve through the depth of the network. As a robustness check against heavy tails, we also compute normalized variants using the median and a 5\% trimmed mean in place of $\overline{\rho}_\ell$ (see left panel of Figure~\ref{fig:llama8b-layerwise-geometry}). Qualitative conclusions did not change for the models explored. We plot the mean-normalized curves by default.

Across all models we observe a similar characteristic growth of the mean norm with depth, often with a sharp increase in the final layers. This norm growth largely drives the scale of the \emph{raw} margin and co-Lipschitz curves. The raw diagnostics are therefore informative about robustness to fixed-scale perturbations (e.g,\ additive noise or quantization), while the normalized versions remove overall scale and probe the intrinsic geometry on the unit sphere. In Figure~\ref{fig:model-layerwise-geometry} we plot the normalized metrics for several models. These reveal an interesting clustering phenomenon between the model families, which we come back to in section~\ref{subsec:model-families}.

\subsubsection{Dependence on sequence length}

\begin{figure*}[t!]
    \centering

    \begin{subfigure}[b]{0.49\textwidth}
        \centering
        \includegraphics[width=\textwidth]{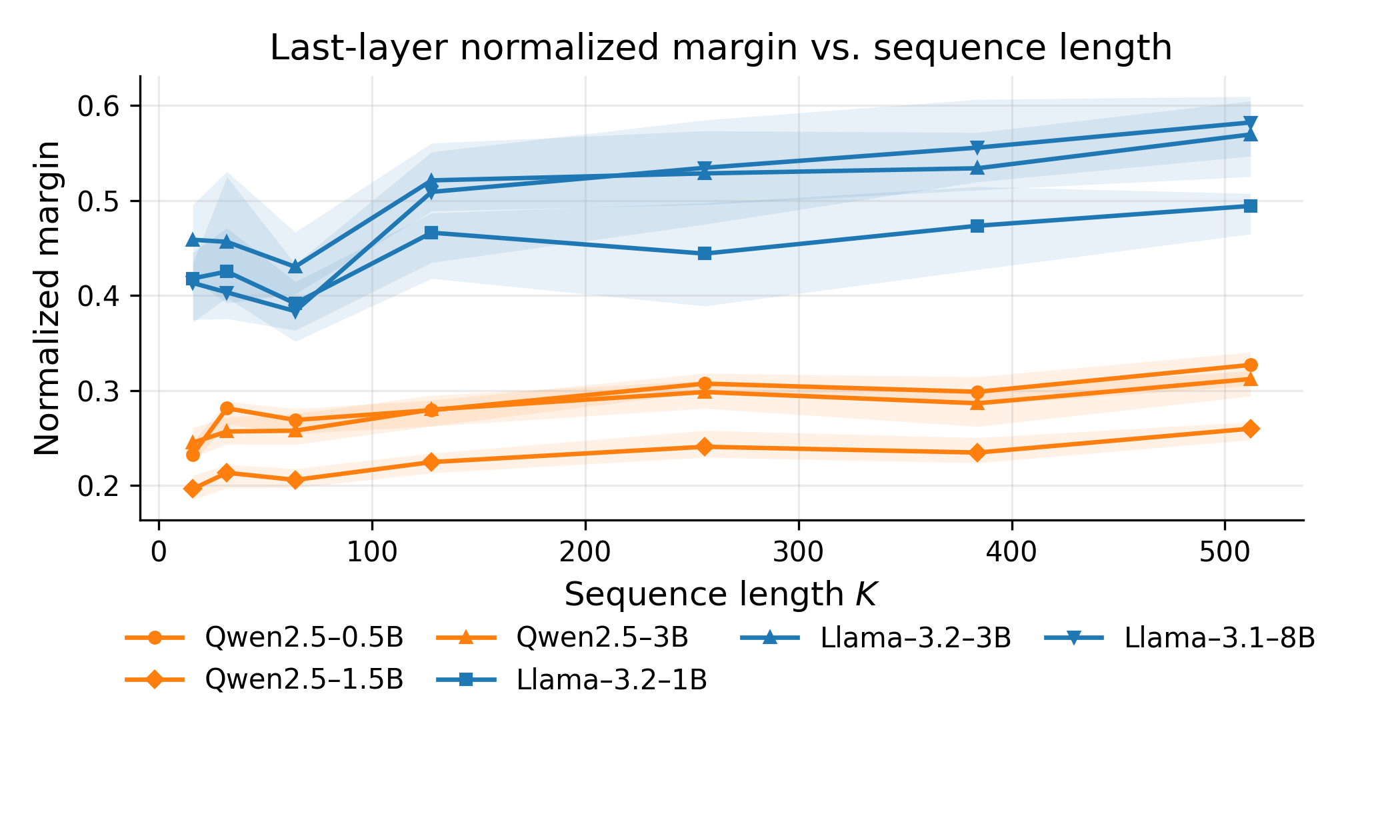}
        \subcaption{Separation margins}
        \label{fig:llama8b-seqlen-margins}
    \end{subfigure}
    \hfill
    \begin{subfigure}[b]{0.49\textwidth}
        \centering
        \includegraphics[width=\textwidth]{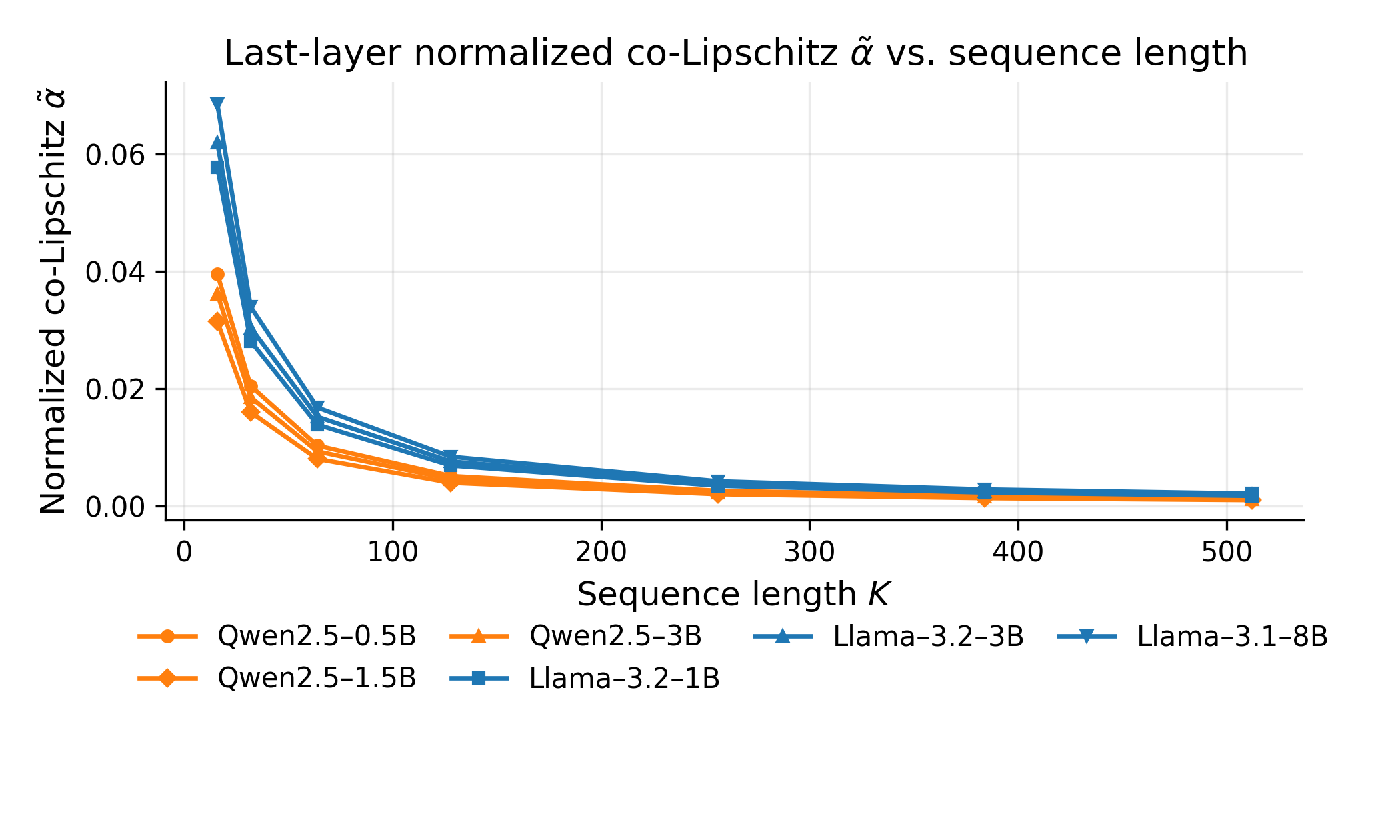}
        \subcaption{Co-Lipschitz estimates}
        \label{fig:llama8b-seqlen-alphas}
    \end{subfigure}

    \caption{Dependence of normalized last-layer margin and co-Lipschitz estimates on sequence length for models in the Qwen (0.5B--3B) and LLaMA-3 (1B--8B) families, $q = 1$\%. Longer contexts tend to reduce the worst-percentile co-Lipschitz constants, indicating more contractive behavior on hard pairs.}
    \label{fig:model-seqlen-geometry}
\end{figure*}

To study how the geometry of the \emph{last layer} depends on the prompt length, we consider a range of context lengths
\[
K \in \{16, 32, 64, 128, 256, 384, 512\}
\]
and, for each $K$, construct a $K$-token prompt set $\mathcal{S}_{\mathrm{exp}}(K)$ (by truncating or filtering as in Section~\ref{subsec:data}) together with the corresponding last-layer states $\{x^{L}_i\}$. For each $K$ we compute:
\begin{itemize}
  \item the last-layer mean norm $\overline{\rho}_{L}(K)$;
  \item the last-layer margin $\widehat{\mathrm{margin}}_{L}^{(q)}(K)$ and co-Lipschitz constant $\widehat{\alpha}_{L}^{(q)}(K)$;
  \item their normalized versions $\widetilde{\mathrm{margin}}_{L}^{(q)}(K)$ and $\widetilde{\alpha}_{L}^{(q)}(K)$;
  \item exact collision counts at the last layer.
\end{itemize}
In Figure~\ref{fig:model-seqlen-geometry} we plot these as functions of $K$, in the spirit of \citet{original_paper}, to see how hard-to-separate pairs and co-Lipschitz behavior are influenced by sequence length.\footnote{In \citet{original_paper} they studied K-dependence of the mean $\ell_2$ norm. Our margin is similar in spirit, but focuses on the hard to separate sequences and therefore address a different question.}

Empirically, we find that normalized margins are relatively stable across sequence lengths, while worst-percentile co-Lipschitz constants tend to decrease for longer contexts, especially in the larger models. This suggests that, for fixed token edits, long-context prompts are represented in a slightly more contractive regime, though still without inducing collisions on the sampled prompt sets.

\subsubsection{Family geometry and model scaling}
\label{subsec:model-families}
To examine how architecture and scale affect the last-layer geometry, we repeat the above procedures across a family of models with increasing parameter counts (several Qwen and LLaMA-3 variants). For a fixed $K$ and $\mathcal{S}_{\mathrm{exp}}(K)$, we compute last-layer margins and co-Lipschitz estimates for each model and consider
\[
\text{params} \;\longmapsto\; \widehat{\alpha}_L^{(q)},\quad \widetilde{\alpha}_L^{(q)},\quad \widehat{\mathrm{margin}}_L^{(q)},\quad \widetilde{\mathrm{margin}}_L^{(q)}.
\]
This probes whether larger models improve intrinsic geometric conditioning (in the sense of co-Lipschitz behavior and separation) on a fixed prompt distribution, and whether different architectural families occupy distinct regions of the diagnostic space.

\begin{figure*}[t!]
    \centering

    \begin{subfigure}[b]{0.49\textwidth}
        \centering
        \includegraphics[width=\textwidth]{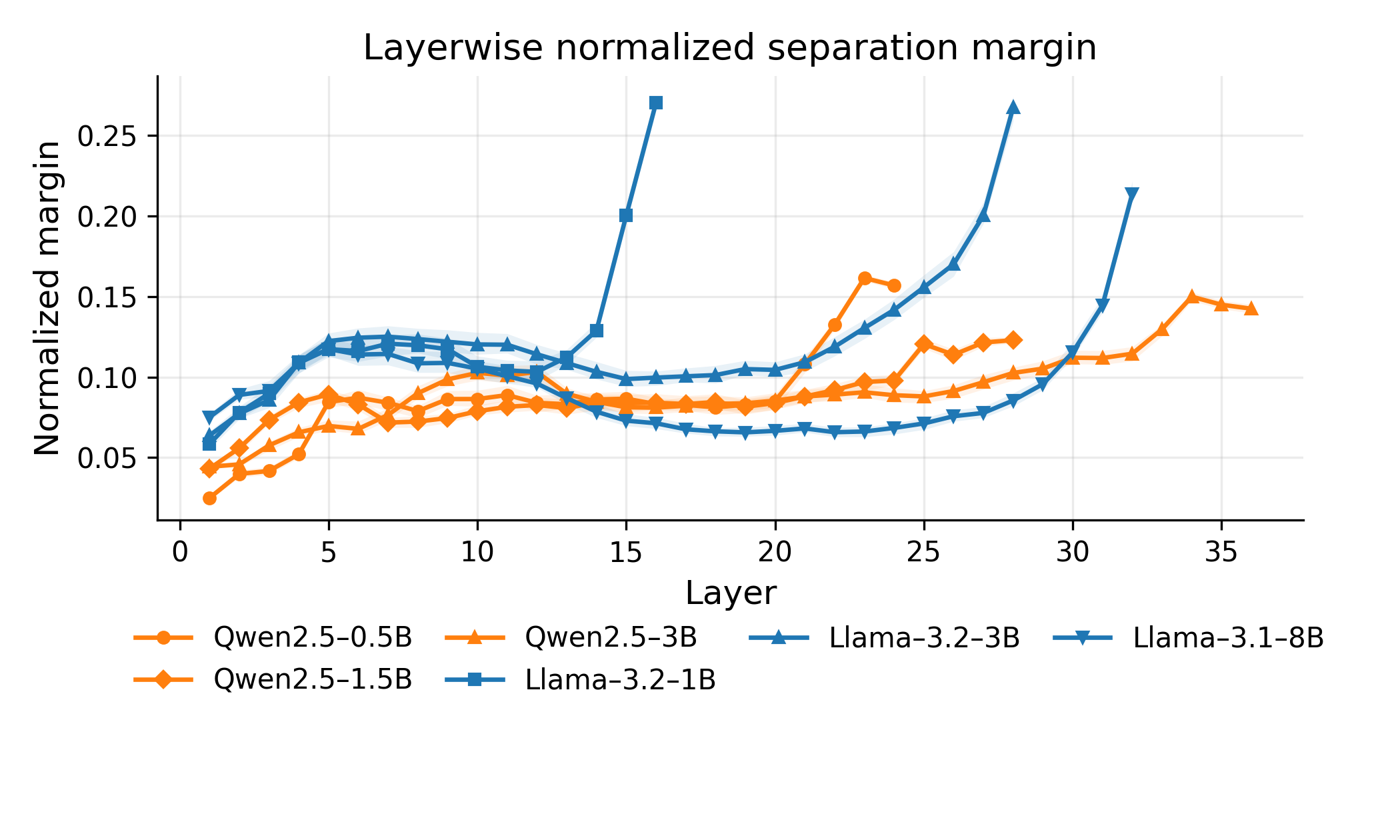}
        \subcaption{Separation margins}
        \label{fig:models-margins}
    \end{subfigure}
    \hfill
    \begin{subfigure}[b]{0.49\textwidth}
        \centering
        \includegraphics[width=\textwidth]{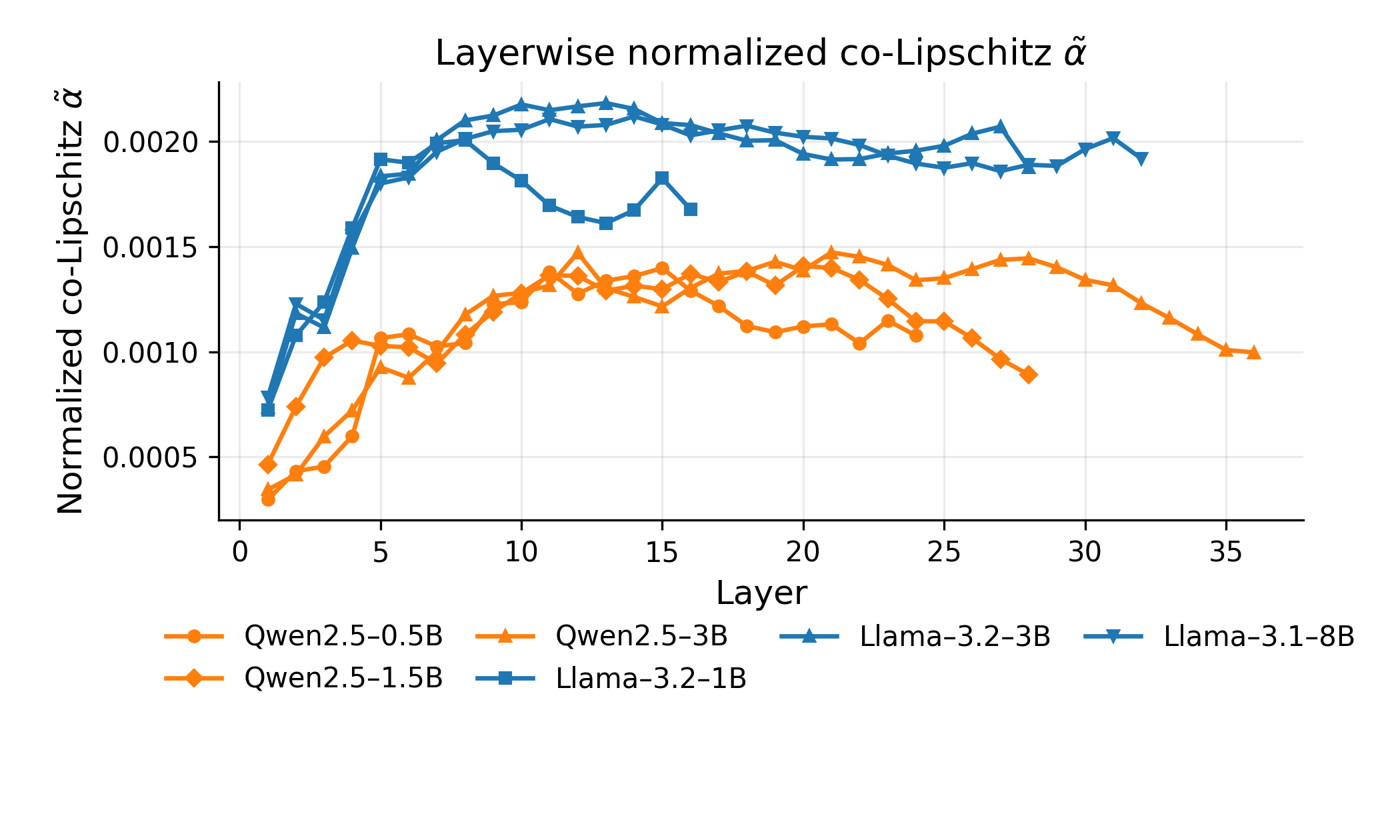}
        \subcaption{Co-Lipschitz estimates}
        \label{fig:models-alphas}
    \end{subfigure}

    \caption{Layerwise normalized separation margins and co-Lipschitz estimates for models in the Qwen (0.5B–3B) and LLaMA-3 (1B–8B) families, $q = 1\%$. Normalization removes the overall growth in representation norms and reveals tight clustering within each model family, together with a consistent offset between families. }
    \label{fig:model-layerwise-geometry}
\end{figure*}

Across all sizes we tested, Qwen and LLaMA families exhibit distinct but internally consistent profiles of normalized margin and co-Lipschitz behavior as a function of depth. Models within the same family cluster tightly, while the two families are separated by a roughly constant offset in these scale-invariant diagnostics (see Figure~\ref{fig:model-layerwise-geometry}). LLaMA models typically have larger normalized margins and co-Lipschitz constants than Qwen at comparable depths, indicating more expansive layerwise geometry and greater worst-case sensitivity to token edits on the unit sphere, whereas Qwen appears systematically more contractive with a flatter mid-depth profile. This suggests that our Lipschitz-inspired metrics capture genuine architectural and training differences rather than merely reflecting parameter count.

Our empirical metrics thus provide an architectural signature: models that share a training recipe and block design cluster tightly in this space, while different families separate. For models that are known to be LLaMA-style with modifications (e.g., Qwen), the geometry confirms that they form their own coherent family, distinct from the original LLaMA series.

\subsubsection{Non-analytic perturbations: uniform activation quantization}
\label{subsec:quantization}

We next probe how fragile the geometric diagnostics are to non-analytic post-hoc distortions by applying simple uniform quantization to hidden states. For a given model and prompt set $\mathcal{S}_{\mathrm{exp}}(K)$, we first run the full-precision network (bf16 weights and activations) and store all last-token hidden states $x^\ell_i \in \mathbb{R}^d$ for every layer $\ell$ and prompt index $i$. 

\paragraph{Quantizer and per-layer scale.}
For each layer $\ell$ we define the \emph{empirical dynamic range}
\begin{equation}
R_\ell \,\equiv\, \max_{i}\,\|x^\ell_i\|_\infty  =  \max_{i,k}\,|x^\ell_{i,k}|,
\end{equation}
computed from these full-precision activations. We then use a \emph{single per-layer symmetric scale} and quantize each coordinate independently by the uniform quantizer
\begin{equation}
Q_b(x)_k \;=\; \delta_{\ell,b}\,\mathrm{round}\!\Big(\frac{x_k}{\delta_{\ell,b}}\Big),\qquad 
\delta_{\ell,b} \;=\; \frac{2R_\ell}{2^b-1}\;\approx\;\frac{2R_\ell}{2^b},
\end{equation}
which maps the interval $[-R_\ell,R_\ell]$ to $2^b$ evenly spaced levels including $0$. This produces quantized hidden states $\tilde x^\ell_i = Q_b(x^\ell_i)$ for $b \in \{8,4\}$, while leaving the underlying model untouched (no retraining, no per-channel scales). This per-layer, max-range scaling is intentionally conservative and may inflate effective step sizes due to outliers. A more sophisticated treatment would involve percentile-clipped ranges or per-channel scales which can reduce distortion. This is however beyond the scope of the current paper.

For each bit-width $b$ we recompute the same layerwise diagnostics on the quantized representations:
\begin{itemize}
  \item the worst-percentile nearest-neighbor separation margin $\widehat{\mathrm{margin}}_\ell^{(q)}$ (raw and normalized by the mean norm at layer $\ell$);
  \item the corresponding co-Lipschitz estimates $\widehat{\alpha}_\ell^{(q)}$ and $\widetilde{\alpha}_\ell^{(q)}$;
  \item exact collision counts, obtained by hashing $\tilde x^\ell_i$ and counting pairs $(i,j)$ with $\tilde x^\ell_i = \tilde x^\ell_j$.
\end{itemize}

\begin{figure*}[t!]
    \centering
    \includegraphics[width=0.75\textwidth]{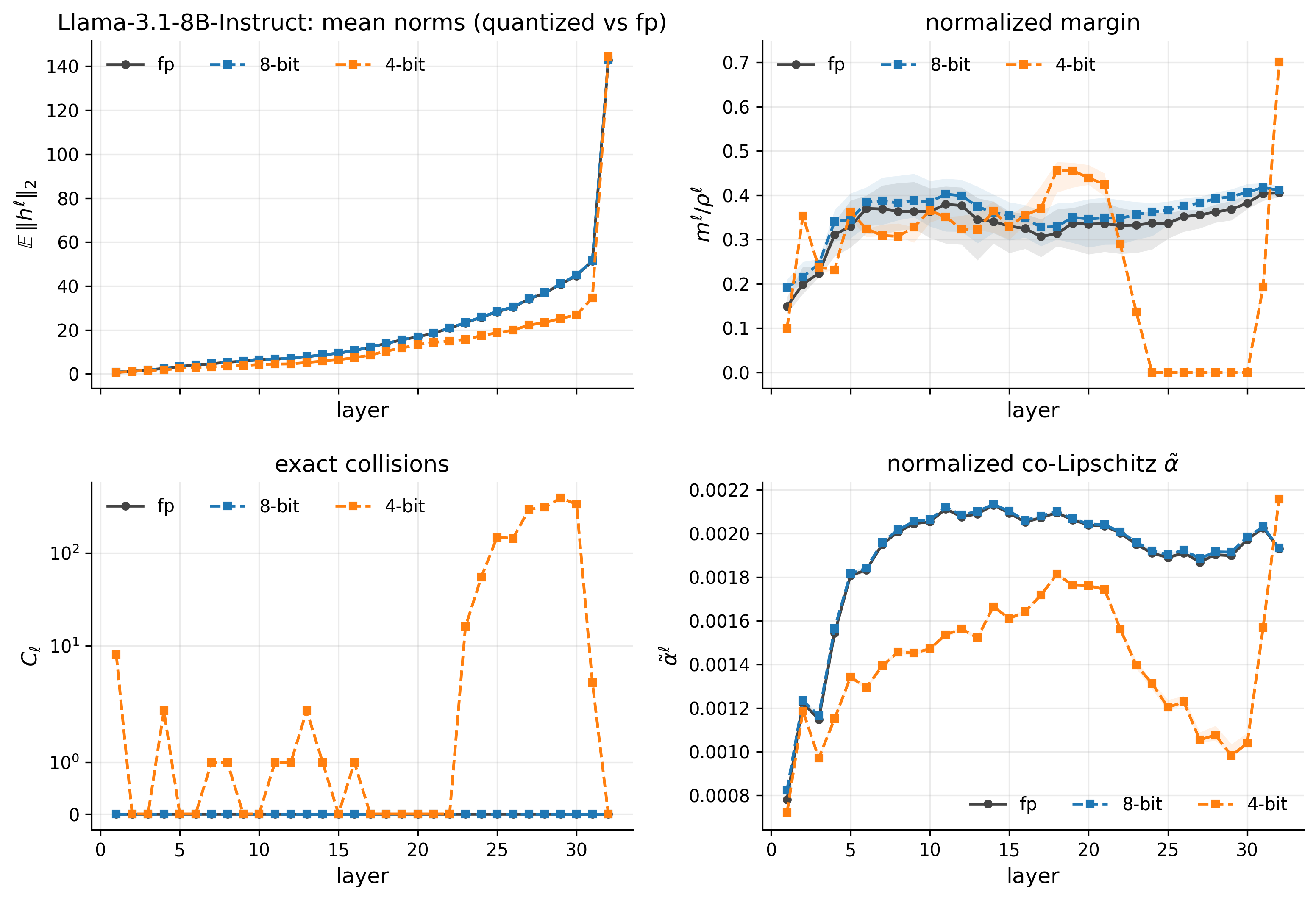}
\caption{Effect of post-hoc uniform per-layer activation quantization (8- and 4-bit) on last-token representations for \texttt{Llama-3.1-8B-Instruct}. Top left: mean $\|h^\ell\|_2$ per layer; Top right: normalized separation margins; Bottom left: exact collision counts per layer $C_\ell$. Bottom right: normalized co-Lipschitz $\tilde \alpha$. Full-precision and 8-bit show no collisions on the sampled prompt set and only mild margin erosion, whereas 4-bit reduces normalized margins more substantially and introduces collisions predominantly in deeper layers, indicating a sharper approach to the discriminant under aggressive discretization.}
\label{fig:llama8b-quantization}
\end{figure*}

We compare the quantized diagnostics to the full-precision baseline ($b=\infty$) as functions of layer and bit-width. Across all models and datasets considered, moderate 8-bit uniform quantization slightly erodes raw margins and co-Lipschitz estimates but produces \emph{no} exact collisions on our finite prompt sets. Under more aggressive 4-bit quantization we observe a clear degradation of raw margins and co-Lipschitz estimates, together with a small but non-negligible number of exact collisions in the quantized representations, concentrated in the deeper layers (Figure~\ref{fig:llama8b-quantization}). In other words, while the full-precision maps $F^\ell_\theta$ remain empirically injective on $\mathcal{S}_{\mathrm{exp}}(K)$, the discretized maps $Q_4 \circ F^\ell_\theta$ can lose injectivity once the resolution becomes sufficiently coarse (e.g.,\ for \texttt{Llama-3.1-8B-Instruct} we observe $1713$ colliding pairs among all prompts at 4-bit, and none at 8-bit). Since $Q_b$ is non-analytic and non-injective, the generic injectivity theorem does not apply to $Q_b \circ F^\ell_\theta$, so collisions for small $b$ are qualitatively consistent with the theory and are expected once the discretization becomes sufficiently coarse.

In this setting we can invoke the robustness result of Theorem~\ref{thm:robust-inj-radius}. Fix a layer $\ell$, a finite prompt set $S \subset \mathcal{S}$, and parameters $\theta$. Let
\begin{equation}
m^\ell_S(\theta)\;=\;\min_{s\neq \tilde s \in S}\,\big\|F^\ell_\theta(s)-F^\ell_\theta(\tilde s)\big\|_2\,,
\end{equation}
denote the (empirically estimated) separation margin on $S$. If the uniform quantizer uses the per-layer step size $\delta_{\ell,b}=2R_\ell/(2^b-1)$ defined above, then by Corollary~\ref{cor:quantization-injectivity} the quantized map $Q_b\!\circ F^\ell_\theta$ remains injective on $S$ whenever
\begin{equation}
\delta_{\ell,b}\;<\;\frac{m^\ell_S(\theta)}{\sqrt{d}}\,.
\end{equation}
In particular, the onset of quantization-induced collisions along layer $\ell$ occurs at a parameter-dependent threshold bitwidth $b^\ell_{\mathrm{crit}}(\theta)$ determined by this inequality. 
Specializing to \texttt{Llama-3.1-8B-Instruct} ($L{=}32$, $d{=}4096$ so $\sqrt{d}{=}64$), the condition becomes
\[
\text{8-bit: } R_\ell \lesssim 2\,m^\ell_S(\theta),
\qquad
\text{4-bit: } R_\ell \lesssim \tfrac{1}{8}\,m^\ell_S(\theta).
\]
Reading $R_\ell$ off from the stored full-precision activations and $m^\ell_S(\theta)$ from Fig.~\ref{fig:llama8b-layerwise-geometry}, the 8-bit inequality is consistent layerwise with the absence of collisions, whereas the 4-bit inequality fails in deeper layers—precisely where collisions appear—aligning the theoretical safety condition with the observed transition. Our approximate diagnostics provide a notion of proximity to this threshold; a finer analysis of the transition itself is left for future work.

\begin{figure*}[t!]
    \centering
    \includegraphics[width=\textwidth]{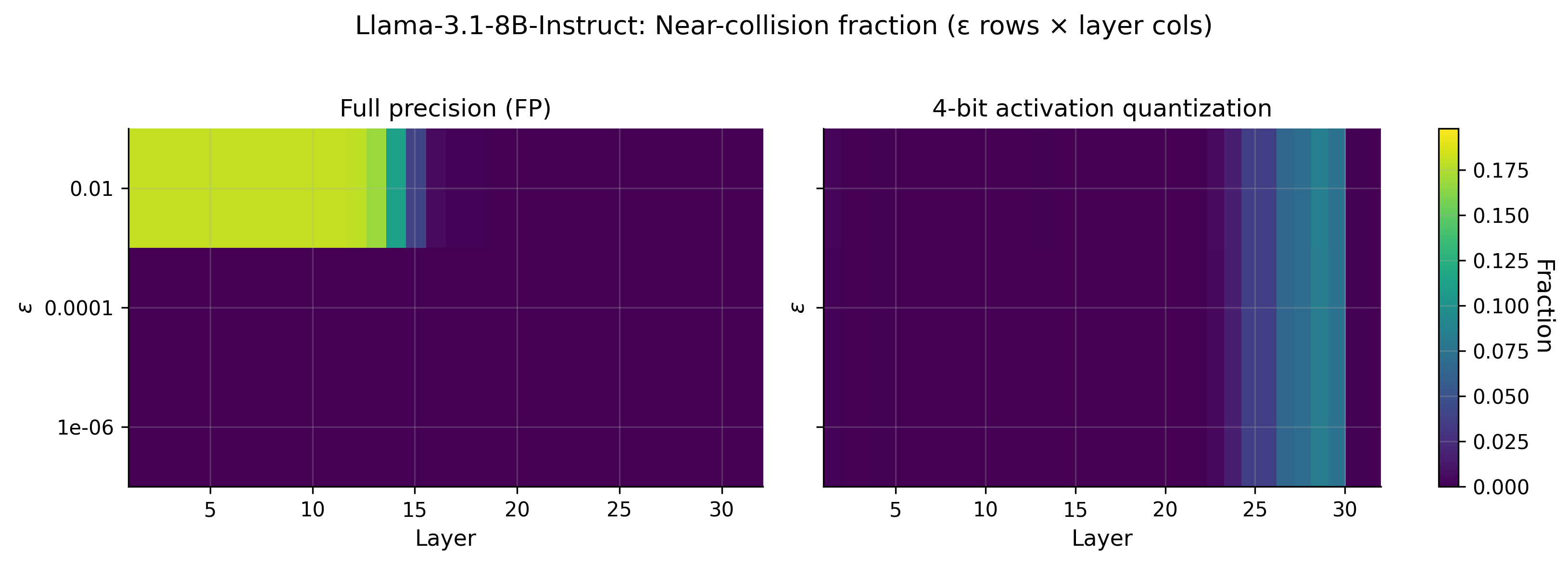}
    \caption{Near-collision fraction by layer for \texttt{Llama-3.1-8B-Instruct}. Each heatmap shows the fraction of prompt pairs whose last-token representations are within a tolerance $\varepsilon$ (rows) at each layer (columns). Left (FP): near-collisions appear only in early layers and only at the loose tolerance $\varepsilon=10^{-2}$; they vanish by mid-depth and are essentially zero for $\varepsilon<10^{-4}$. Right (4-bit activ. quantization): near-collisions persist throughout the network and sharply increase in the final third of layers, indicating depth-amplified discretization effects. Color indicates fraction (shared scale). This pattern is consistent with shrinking safety margins under 4-bit quantization, while full precision remains well separated except at very shallow layers.}
\label{fig:llama8b-eps-sweeps}
\end{figure*}

\paragraph{Near-collision structure under quantization.} To complement Figure~\ref{fig:llama8b-quantization}’s margin and collision counts, Figure~\ref{fig:llama8b-eps-sweeps} shows the fraction of prompt pairs whose last-token representations lie within a tolerance $\varepsilon\in\{10^{-6}, 10^{-4}, 10^{-2}\}$ at each layer, comparing full precision (left) and 4-bit activation quantization (right). Full precision exhibits appreciable near-collisions only at the loosest tolerance and only in shallow layers, whereas 4-bit quantization sustains elevated near-collision rates across depth with a sharp rise in the final third—consistent with the shrinking safety margins and the onset of collisions seen in Figure~\ref{fig:llama8b-quantization}.

\subsubsection{Training trajectories}

For a smaller Transformer that can be trained from scratch or fine-tuned cheaply, we also study the evolution of geometry along a training trajectory. Starting from a random or pretrained initialization, we save checkpoints at regular intervals and, at each checkpoint, compute last-layer metrics on a fixed prompt set $\mathcal{S}_{\mathrm{exp}}(K)$. Tracking
\[
\widehat{\mathrm{margin}}_L^{(q)}(t),\quad \widetilde{\mathrm{margin}}_L^{(q)}(t),\quad
\widehat{\alpha}_L^{(q)}(t),\quad \widetilde{\alpha}_L^{(q)}(t)
\]
as functions of training step $t$ gives an empirical test of the persistence-of-injectivity picture from Theorem~\ref{theorem:persistence}. For the GPT-2 model we trained, we do not observe any systematic drift in these metrics over the course of training: both normalized margin and normalized co-Lipschitz remain within a narrow band around their initialization values: \ $\widehat{\mathrm{margin}}_L^{(q)} \approx 0.2 \pm 0.04$ and $\widehat{\alpha}_L^{(q)} \approx 0.002 \pm 0.0003$.

\subsection{Discussion of metrics and alternative geometries}
\label{subsec:discussion}

The two main diagnostics, $\widehat{\mathrm{margin}}_\ell^{(q)}$ and $\widehat{\alpha}_\ell^{(q)}$, have complementary interpretations. The margin captures how tightly packed the hardest-to-separate pairs of representations are, while the co-Lipschitz constant measures how much the representation moves per unit of Hamming distance in the most contractive cases. In particular, with Hamming distance on $V^K$,
\[
\widehat{\alpha}_\ell^{(q)} \;\approx\; \text{``minimum representation movement per token edit''}
\]
up to a worst-percentile relaxation. This aligns naturally with the injectivity theorem, which is formulated on the discrete prompt set $\mathcal S$.

In principle, one could use these diagnostics to compare pretrained Transformers against simple baselines such as random linear embeddings or untrained Transformer stacks of the same width, in order to isolate architectural versus training effects on “distance to collision.” We leave such systematic baseline comparisons to future work, and here focus on pretrained models and a small trained-from-scratch GPT-2, for which we can relate our empirical findings more directly to the analytic injectivity picture.

The normalized versions $\widetilde{\mathrm{margin}}_\ell^{(q)}$ and $\widetilde{\alpha}_\ell^{(q)}$ factor out global scale and are best viewed as scale-invariant condition numbers: they quantify how well-separated and how co-Lipschitz the layerwise maps are when restricted to the unit sphere. In contrast, the raw quantities retain information about norm inflation and are directly relevant for robustness to fixed-scale noise, quantization, and other perturbations that do not rescale representations.

While we focus on Hamming distance on token sequences, one could also consider alternative input geometries. For example, let $E:V^K\to\mathbb{R}^p$ denote a learned embedding map (e.g.,\ the model’s own embedding layer applied to the prompt), and define
\[
d_E(s,\tilde s) = \|E(s)-E(\tilde s)\|_2.
\]
Co-Lipschitz estimates based on $d_E$ probe the Lipschitz behavior of the composite $F^\ell_\theta\circ E^{-1}$ on the image of $E$, and thus the compatibility between the geometry of the input embedding space and that of downstream representations. Exploring such alternative input metrics is an interesting direction for future work, but for the purposes of generic injectivity and discrete invertibility, the Hamming geometry on $\mathcal S$ is the most natural choice.

\subsection{Results}
Across all tested models and configurations, we find a consistent picture. Raw margins and co-Lipschitz estimates typically grow with depth, driven largely by norm inflation, while their normalized counterparts remain roughly flat as functions of layer index. Worst-percentile co-Lipschitz constants decrease with sequence length, indicating more contractive behavior on long contexts. Moderate post-hoc activation quantization (8-bit) erodes raw margins and co-Lipschitz estimates but does not introduce collisions on our sampled prompts, whereas more aggressive 4-bit quantization can induce collisions in the quantized representations, concentrated in deeper layers. Finally, along the pretraining trajectory of a small GPT-2 model, both normalized margin and normalized co-Lipschitz remain essentially stable, providing empirical support for the persistence-of-injectivity picture suggested by our analytic results.

\section{Discussion and Conclusion}
Our analytic results show that, under real-analytic assumptions on the architecture and mild non-singularity conditions on the optimizer, decoder-only Transformers are generically injective on any fixed finite prompt set in the continuous parameter idealization, and that this property is almost surely preserved along smooth training trajectories. Our empirical study complements this by quantifying \emph{how far} a given model appears to be from losing injectivity, via simple layerwise diagnostics based on separation margins and co-Lipschitz constants.

From a practical standpoint, these diagnostics offer a lightweight tool for practitioners: given a pretrained model and a task-specific prompt distribution, one can estimate layerwise margins and co-Lipschitz constants to assess (i) whether post-hoc activation quantization at a given bitwidth is likely to introduce representation collisions on that distribution, and (ii) whether architectural or training modifications move the model towards a more or less contractive regime on hard prompt pairs. While we stop short of prescribing concrete deployment thresholds, our results suggest that such geometric summaries can act as useful fingerprints when comparing architectures, quantization schemes, or training recipes from the perspective of potential invertibility.

Across LLaMA and Qwen families and a range of sizes, these scale-invariant diagnostics yield consistent architectural signatures: models that share a block design and training recipe cluster tightly in the space of normalized margins and co-Lipschitz constants, while different families are separated by systematic shifts. We also observe that longer contexts tend to push models into a more contractive regime on the hardest prompt pairs, and that non-analytic perturbations such as uniform activation quantization can dramatically affect raw margins without necessarily causing collisions on the sampled prompts. We emphasize that we do not attempt to directly correlate these geometric diagnostics with the success probability of concrete inversion attacks or reconstruction procedures. Instead, margins and co-Lipschitz estimates should be viewed as preconditions for invertibility: they quantify how close a model is, on a given finite prompt set, to violating injectivity or to operating in a highly contractive regime. Establishing quantitative links between these diagnostics and the empirical performance of supervised inverters or privacy attacks is an interesting direction for future work.

Taken together, these findings suggest a unified picture. Analytically, Transformers are generically and persistently injective in this finite-set, continuous-parameter sense; empirically, they behave like smooth, roughly scale-invariant embeddings of the prompt set, whose practical invertibility can be probed via simple geometric diagnostics. In this sense, our work complements the exact inversion results of \cite{original_paper}: instead of proposing a new inverse algorithm, we provide a quantitative, geometry-based notion of “distance to collision’’ that can be tracked across architectures, depths, context lengths, and non-analytic perturbations. We view extending these diagnostics to other architectures, modalities, and training regimes as a promising direction for future work.

\bibliography{bibliography}
\bibliographystyle{tmlr} 

\end{document}